\documentclass{article} 
\PassOptionsToPackage{table}{xcolor}
\usepackage{iclr2027_conference,times}

\usepackage{amsmath}
\usepackage{amssymb}
\usepackage{amsfonts}

\usepackage{amsmath,amsfonts,bm}

\def\eqref#1{equation~\ref{#1}}

\def\1{\bm{1}}

\DeclareMathAlphabet{\mathsfit}{\encodingdefault}{\sfdefault}{m}{sl}
\SetMathAlphabet{\mathsfit}{bold}{\encodingdefault}{\sfdefault}{bx}{n}

\usepackage{hyperref}
\usepackage{url}
\usepackage{booktabs}
\usepackage{graphicx}
\usepackage{caption}
\usepackage{wrapfig}
\usepackage{float}
\usepackage{algorithm}
\usepackage{titletoc}
\usepackage{algpseudocode}
\usepackage{wrapfig}
\usepackage{xcolor}
\definecolor{MultimodalityGreen}{HTML}{55A868}
\definecolor{DensityGold}{HTML}{CCB974}
\definecolor{StateEvolutionRed}{HTML}{C44E52}
\usepackage{graphicx}
  \usepackage{verbatim}
  \IfFileExists{algorithm.sty}{\usepackage{algorithm}}{}
  \IfFileExists{algpseudocode.sty}{\usepackage{algpseudocode}}{%
    \newenvironment{algorithm}[1][]{\par\medskip\noindent\renewcommand{\caption}[2][]{}\textbf{Algorithm (omitted: algpseudocode unavailable in build).}\par}{\par\medskip}%
     }
  \usepackage{textcomp}

\usepackage{array}
\usepackage{xcolor}

\definecolor{CategoryAvgBlue}{RGB}{242,248,255}
\definecolor{TotalAvgPink}{RGB}{255,244,247}

\title{DSV-Mem: Evaluating multimodal memory in professional workflows for MLLM agents}

\iclrfinalcopy

\author{%
\textbf{Jike Zhong}$^{1,2}$%
\thanks{Equal contribution.}\kern0.4em%
\thanks{Work done during an internship at Amazon AGI.}
\quad \textbf{Ritwick Chaudhry}$^{1}$\footnotemark[1]
\quad \textbf{Xuanbai Chen}$^{1}$ \quad Tianchen Zhao$^{1}$ \\\textbf{Linghan Xu}$^{1}$ \quad
\textbf{Yifan Xing}$^{1}$ \quad \textbf{Nishant Sankaran}$^{1}$ \\
$^{1}$Amazon AGI
\qquad $^{2}$University of Southern California
}

\begin{document}

\maketitle

\begin{abstract}
Conversational MLLM agents are increasingly expected to assist in professional workflows: from AI research and engineering design to product management and business operations. Despite this growing need, this capability has not been well explored: existing benchmarks largely focus on informal, everyday interactions and personal-life scenarios that typically feature photographic natural images, isolated, static artifacts, and recall-oriented questions. In contrast, professional scenarios often involve structured, information-heavy artifacts that undergo frequent revisions and authority updates, as well as compositional queries that require reconciling many artifact versions in memory while tracking state precisely. To address these challenges, we introduce \textbf{DSV-Mem}, a carefully curated benchmark for evaluating \textbf{D}ense \textbf{S}tateful \textbf{V}isual \textbf{Mem}ory. DSV-Mem comprises a diverse suite of expert-reviewed scenarios and 1,000 questions across five user-oriented categories (Current State, Past State, Derived State, Change History, and Conflict/Refusal). Moreover, a Hartley-inspired criterion is introduced to favor questions with broader visual-evidence inspection demands. To support further study, we introduce a generation harness that produces high-quality, domain-specific evaluation suites by decoupling
state-transition synthesis from conversation filling. Rigorous evaluation over 27 configurations spanning frontier, open-weight models and popular memory management methods reveals that even the strongest baseline scores below $45\%$ on DSV-Mem. Comprehensive analysis surfaces several important findings: 1) multimodality and information density both contribute to difficulty, but \emph{state evolution}, particularly the number of governing updates, is the dominant tested factor. Raw conversation/haystack length, OCR and arithmetic are not the primary bottlenecks; 2) models frequently fail to verify user premises against prior state updates before answering; 3) popular interventions such as increased reasoning effort and memory management methods yield limited gains, whereas state-aware designs prove more effective. The benchmark \& code will be publicly released. 
\end{abstract}

\begin{figure}
\centering
\includegraphics[width=\textwidth]{"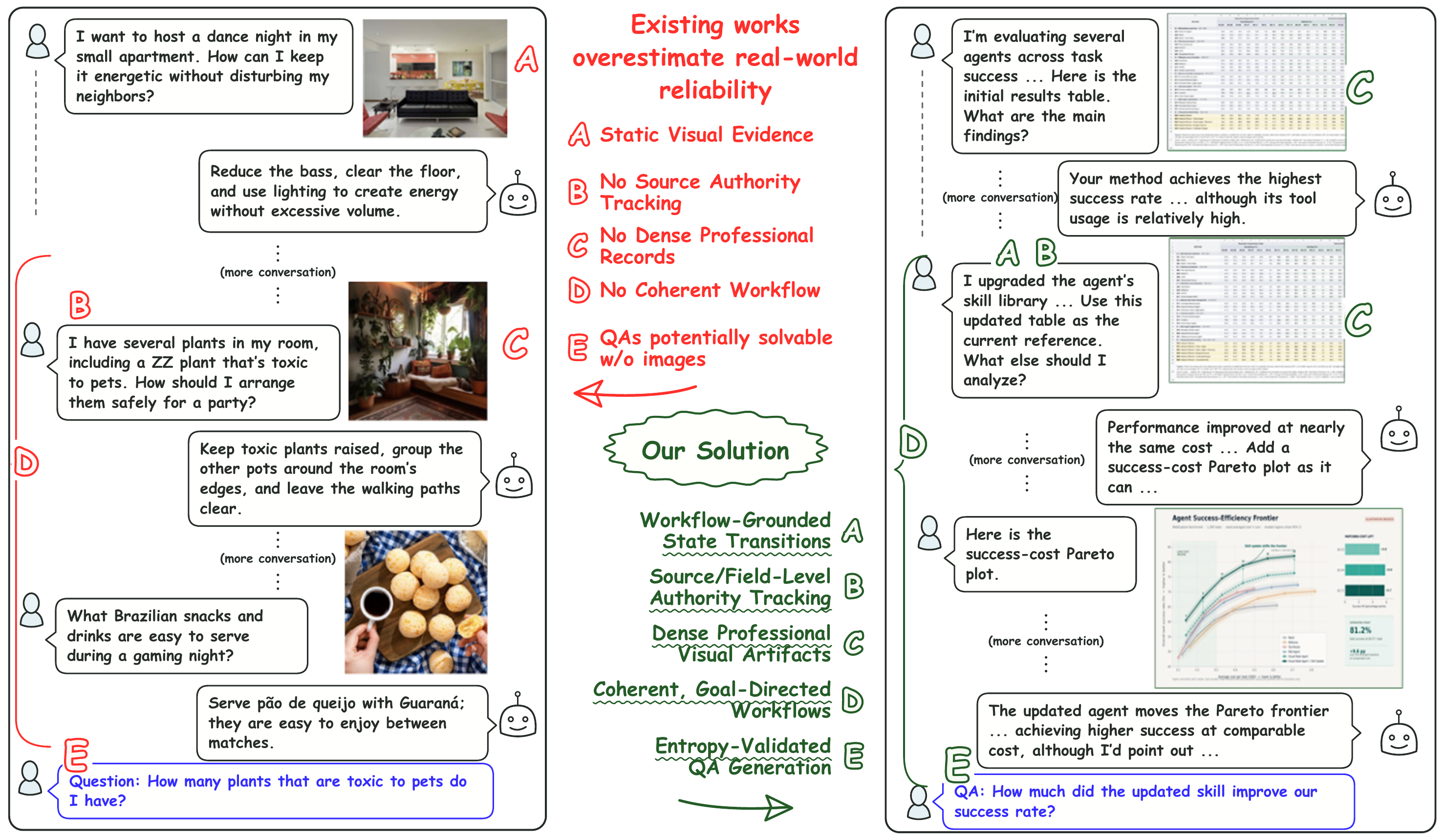"}
\caption{\textbf{Comparison between existing benchmarks and DSV-Mem.}
Existing benchmarks (left) pose QAs over casual conversations with sparse, static photos, potentially overestimating real-world reliability (\textcolor{red!75!black}{A--E}).
DSV-Mem addresses these limitations through goal-directed workflows over dense, evolving professional artifacts (right), with QAs requiring authoritative-state tracking across versions (\textcolor{green!45!black}{A--E}).
Matching letters link each point to the illustrated examples on both panels.}
\label{fig:comparison}
\vspace{-10pt}
\end{figure}


\section{Introduction}
\label{sec:intro}
Conversational MLLM agents are increasingly deployed in extended professional workflows beyond isolated everyday tasks \citep{hu2025osagents, xi2023rise, wang2025uitars}. For instance, designers generate and iteratively refine complex campaign graphics, while ML researchers repeatedly analyze evolving tables and plots, and revise paper drafts as experiments progress. Unlike everyday conversations, these scenarios often exhibit distinctive features: \textbf{1) information-heavy visuals:} they center on structured, data-heavy artifacts such as dense worksheets, dashboards, or intricate graphic designs rather than ordinary photographic scenes; \textbf{2) statefulness:} uploaded artifacts frequently undergo revisions, scope changes, and authority updates; and \textbf{3) analytical queries:} users go beyond single-fact questions to compositional ones that compare, aggregate, or derive quantities across visuals and revisions. Meeting these demands requires robust stateful multimodal memory, such as precise state tracking that reconciles evidence across many versions, and assessing such capabilities is essential for measuring progress toward reliable professional MLLM agents.

Despite this growing need, existing efforts to evaluate multimodal memory largely focus on informal, everyday interactions and personal-life scenarios, featuring photographic visuals \citep{maharana2024locomo, bei2026memgallery, ren2026memlens, guo2026memeye}, static artifacts \citep{maharana2024locomo, bei2026memgallery, ren2026memlens, huang2026m3exam, zhu2026h2hmem}, or recall-oriented questions \citep{wu2026longmemevalv2, liu2026worldmemarena}, rather than the dense visuals, stateful artifacts, and analytical queries that professional workflows demand. Moreover, our careful analysis shows many of them average fewer than three evidence images per question, often one or none~\citep{bei2026memgallery}, and may be largely saturated for frontier systems: Gemini-3.7-Flash already exceeds 90\% on LoCoMo~\citep{maharana2024locomo}. Most importantly, none of these settings jointly capture the properties and complexity of real-world workflows encountered in practice (\autoref{fig:comparison}).


To bridge this gap, we introduce DSV-Mem (\textbf{D}ense \textbf{S}tateful \textbf{V}isual \textbf{Mem}ory), a comprehensive benchmark designed around the core properties above to assess MLLM agent memory in professional workflows. DSV-Mem features realistic scenarios with expert-reviewed dense artifacts, extended and topic-centric conversations with diverse stateful evolutions, and user-oriented categories (Current State, Past State, Derived State, Change History, and Conflict/Refusal) covering diverse real-world requests. 

However, curating such a benchmark is highly nontrivial: Without explicit state propagation control, directly prompting an LLM/agent often yields repetitive dialogues with homogeneous or corrupted state transitions. These degraded trajectories in turn produce shallow QAs with incorrect, unverifiable labels. Conversely, manual creation limits controlled modifications, making systematic study of the underlying factors difficult. To this end, we develop a generation harness that produces high-quality conversations requiring only modest curation effort. Our key design principles are 1) the decoupling of state-trajectory construction from surface conversation realization, and 2) a Hartley-inspired criterion for cross-visual QA construction, with 7.78 candidate visual artifacts per QA in DSV-Mem (geometric mean), substantially higher than similar benchmarks (\autoref{tab:evidence_entropy}).

We conduct rigorous evaluation of 27 configurations spanning advanced frontier and open MLLMs and popular memory management methods. DSV-Mem remains hard for all evaluated baselines: the best-performing baselines among our evaluations, Gemini-3.7-Flash, achieves below 45\% with over 80\% of the errors concentrated on state reconstruction rather than computation or OCR. Beyond this, a large memory capability gap exists between the best \emph{compact} open-weight systems and frontier MLLMs. Common interventions offer little: increased reasoning effort shows limited gains, and existing memory management solutions or simple supervised fine-tuning (SFT) provide limited gains. Moreover, a controlled ablation shows all three properties matter but statefulness dominates: removing statefulness improves matched-category accuracy by
43.2 pp, versus 12.6 pp for density reduction and 8.2 pp
for visual-to-text conversion. Based on these insights, we explore a lightweight state-aware inference-time algorithm that proves effective ($+13.6-26\%$ overall), yet further improvement may require efforts beyond simple scaffolding or SFT. Our contributions are threefold: \begin{itemize}
    \item \textbf{A comprehensive benchmark:} We identify the lack of assessment tools for conversational MLLM agents in professional workflows and introduce DSV-Mem, a benchmark built around their defining features: long conversations over dense, stateful artifacts and analytical queries. 
    \item \textbf{A generation harness:} We introduce an automated harness that synthesizes coherent, domain-specific workflow dialogues with diverse state updates, requiring minimal post-hoc curation. This paves the way for scalable future study.   
    \item \textbf{Evaluations and findings:} We evaluate 27 configurations, conduct in-depth analysis, and share important insights that could inform better design of MLLM conversational agents in the future.    
\end{itemize}

\begin{figure}
\centering
\includegraphics[width=\textwidth]{"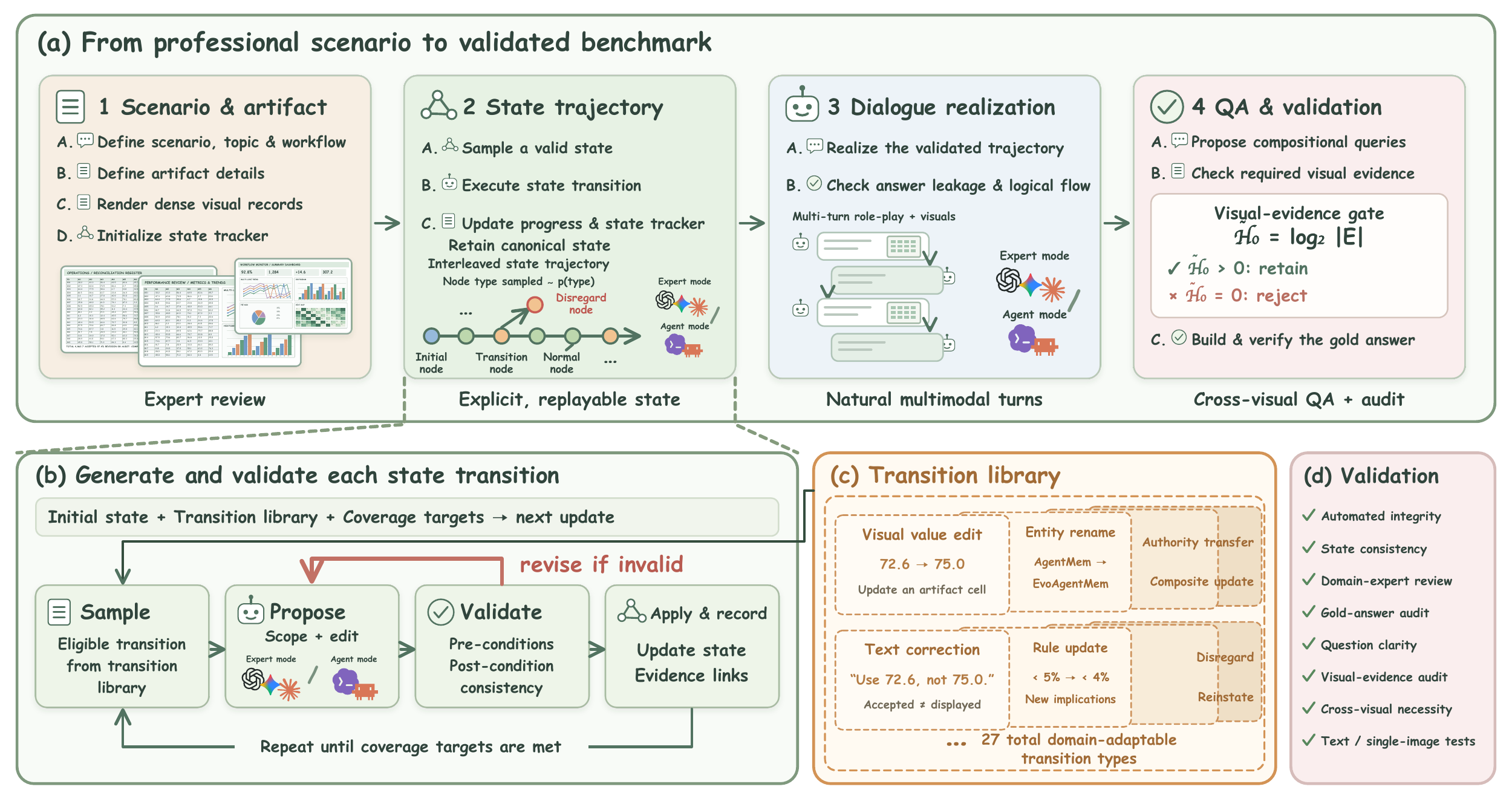"}
\caption{\textbf{DSV-Mem benchmark construction pipeline.}
\textbf{(a)} Generation separates explicit state-trajectory construction from natural dialogue realization, followed by compositional QA construction with cross-visual evidence requirements.
\textbf{(b)} The propose--validate--revise loop constructs transitions while preserving canonical state and evidence links.
\textbf{(c)} A reusable library provides 27 domain-adaptable transition types.
\textbf{(d)} Extensive automated checks and human audits.}
\label{fig:overall_pipeline}
\vspace{-10pt}
\end{figure}

\section{Related Work}

\paragraph{MLLM memory benchmarks.}
LoCoMo, LongMemEval, and EverMemBench evaluate conversational recall, temporal reasoning, and knowledge updates, but do not center on reconstructing fine-grained state across revisions of dense visual artifacts \citep{maharana2024locomo,wu2025longmemeval,hu2026evermembench}.
Mem-Gallery, MemLens, and M$^3$Exam evaluate multimodal histories and heterogeneous evidence, while H2HMem studies human--human interactions; these settings do not focus on cumulative artifact revisions and authority changes within sustained professional workflows \citep{bei2026memgallery,ren2026memlens,huang2026m3exam,zhu2026h2hmem}.
LongMemEval-V2 and WorldMemArena include dynamic state tracking, but emphasize reusable environment knowledge and action--world memory, respectively, rather than compositional analysis of dense, repeatedly revised records \citep{wu2026longmemevalv2,liu2026worldmemarena}.
MemEye evaluates fine-grained visual evidence and evolutionary synthesis, but emphasizes visual granularity and reasoning depth across life scenarios \citep{guo2026memeye}.
DSV-Mem instead centers on the \textbf{interaction between visual density and state evolution}: compositional queries require reconciling field-level updates and source authority across revisions of dense professional artifacts, rather than recalling individual state changes in isolation.

\paragraph{MLLM memory-management methods.}
Generative Agents and MemoryBank emphasize reflection and personalized recall, while MemGPT and MemoryOS manage hierarchical memory, prioritizing conversational continuity rather than reconstruction of evolving visual records \citep{park2023generativeagents,zhong2023memorybank,packer2024memgpt,kang2025memoryos}.
LongMem retrieves cached text, HippoRAG links knowledge for multi-hop retrieval, and A-Mem evolves interconnected notes, emphasizing contextual access and semantic organization rather than explicit field-level authority across artifact versions \citep{wang2023longmem,gutierrez2025hipporag,xu2025amem}.
Mem0 extracts and updates salient facts, while SimpleMem and LightMem compress and consolidate interactions; such compression poses challenges for professional workflows, where minor details may govern later state-dependent queries \citep{chhikara2025mem0,liu2026simplemem,fang2026lightmem}.
MIRIX and MemVerse support multimodal memory, but retaining evidence leaves open how to reconcile successive revisions and source authority \citep{wang2025mirix,liu2026memverse}.
DSV-Mem tests these demands: all four evaluated integrations underperform their baselines, highlighting the challenge of \textbf{preserving and reconciling dense, state-bearing details}.

\section{DSV-Mem}
\begin{wrapfigure}{r}{0.34\textwidth} 
\vspace{-70pt}
\centering
\includegraphics[width=\linewidth]{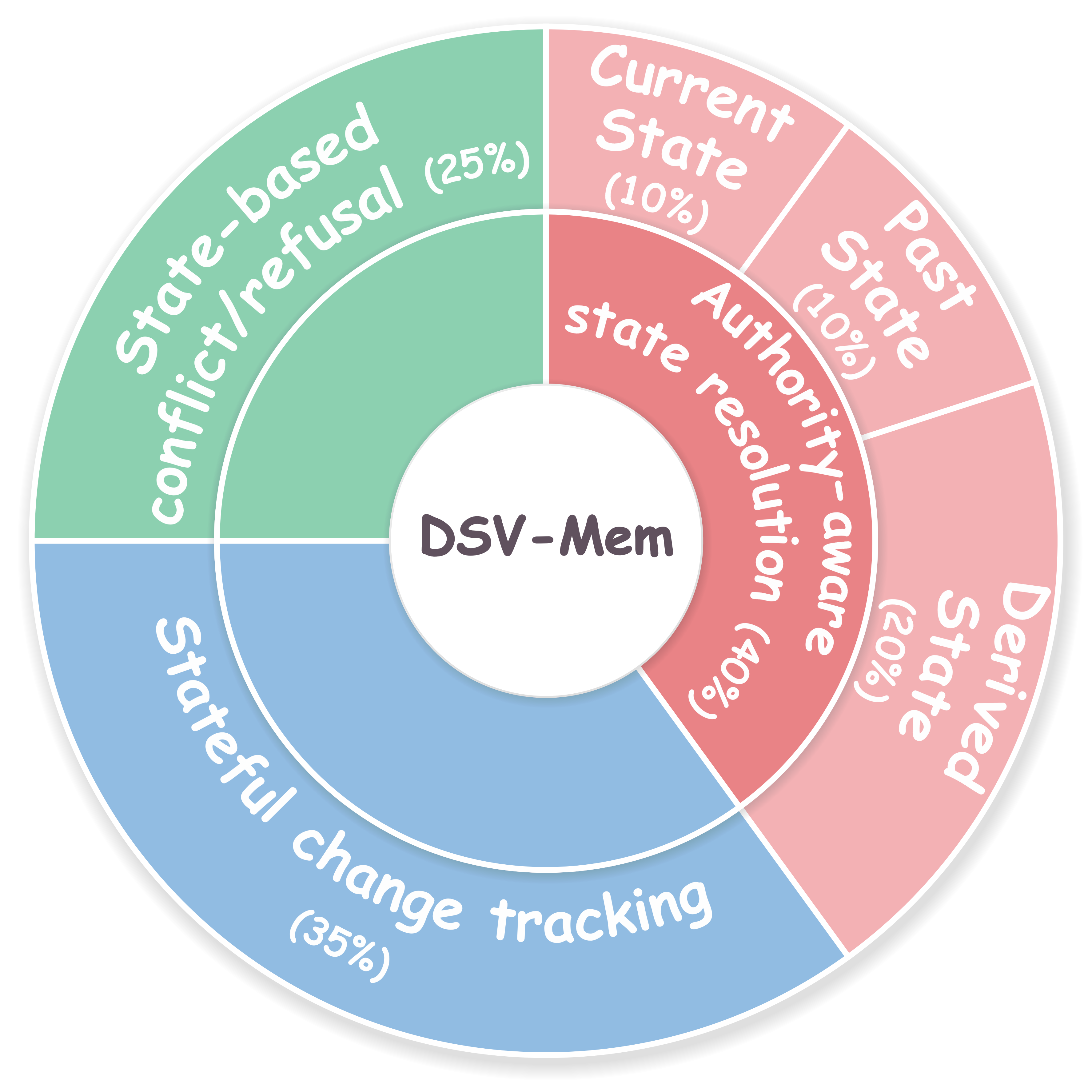} 
\caption{DSV-Mem category breakdown.}
\vspace{-25pt}
\label{fig:cat_pie}
\end{wrapfigure}
\subsection{Benchmark overview}
DSV-Mem comprises 1,000 carefully curated compositional user queries across 10 professional
scenarios (\autoref{app:scenario_overview}), each grounded in a long, coherent conversation with structured, information-heavy artifacts that undergo successive revisions and diverse authority updates.
It extensively evaluates whether MLLM agents can \textbf{1)} track authoritative
multimodal state, \textbf{2)} retain and reason over update history and evolution trajectory, and
\textbf{3)} detect conflicts with maintained state and refuse unsupported requests. On average, each scenario features 212 conversation turns with 24 dense
visual artifacts and up to 50k tokens.

\paragraph{Question categories.}
Questions span five capability axes, grouped into three families (illustrated in \autoref{fig:cat_pie}) that mirror the capabilities mentioned above; counts in parentheses. 
\emph{Authority-aware state resolution} covers \textbf{Current State} (100), identifying the latest authoritative value; \textbf{Past State} (100), 
recovering the state at a specified point in the conversation; 
and \textbf{Derived State} (200), filtering, comparing, or computing over reconstructed state.
\emph{Stateful change tracking} consists of \textbf{Change History} (350), recovering the sequence or frequency of updates.
\emph{State-based conflict/refusal} comprises \textbf{Conflict/Refusal} (250), detecting false premises, unresolved conflicts, or insufficient evidence.
More details are in \autoref{app: question categories}.

\subsection{Benchmark Construction and Validation}

\paragraph{Stateful conversation generation.}
Rigorous evaluation of dense stateful visual memory requires conversation trajectories that jointly satisfy \textbf{three constraints}: \textbf{(i)} domain-faithful dense visual artifacts, \textbf{(ii)} diverse and explicit state transitions with verifiable ground truth, and \textbf{(iii)} coherent long-form interactions. However, directly prompting an LLM to generate complete trajectories often
produces repetitive dialogue, poorly motivated shifts in the workflow, and limited transition diversity. More fundamentally, unlike static settings, evolving scenarios require accepted state to be explicitly propagated and validated across all updates to prevent
inconsistent evidence and QA labels. To overcome these challenges, we \textbf{decouple latent state-trajectory construction from dialogue realization} through a generation harness (\autoref{fig:overall_pipeline}). Starting from an initial artifact and state, the harness samples eligible operations---such as visual-value edits, textual corrections, and authority changes---from a predefined transition library. Each operation is instantiated against the current state, checked for pre- and post-condition consistency, and recorded with its accepted effects and evidence provenance. Two-model role-play then realizes the validated trajectory as natural dialogue, interleaving updates with ordinary interaction. The resulting dialogue is then checked for coherence, state consistency, and answer leakage. This decoupled design enables controlled, diverse trajectories and coherent dialogue, keeps QAs traceable to validated states, and requires only light post-hoc review. Full details are specified in \autoref{app:transition_library} and \autoref{app: detailed pipeline overview}.

\begin{figure}
\centering
\includegraphics[width=\textwidth]{"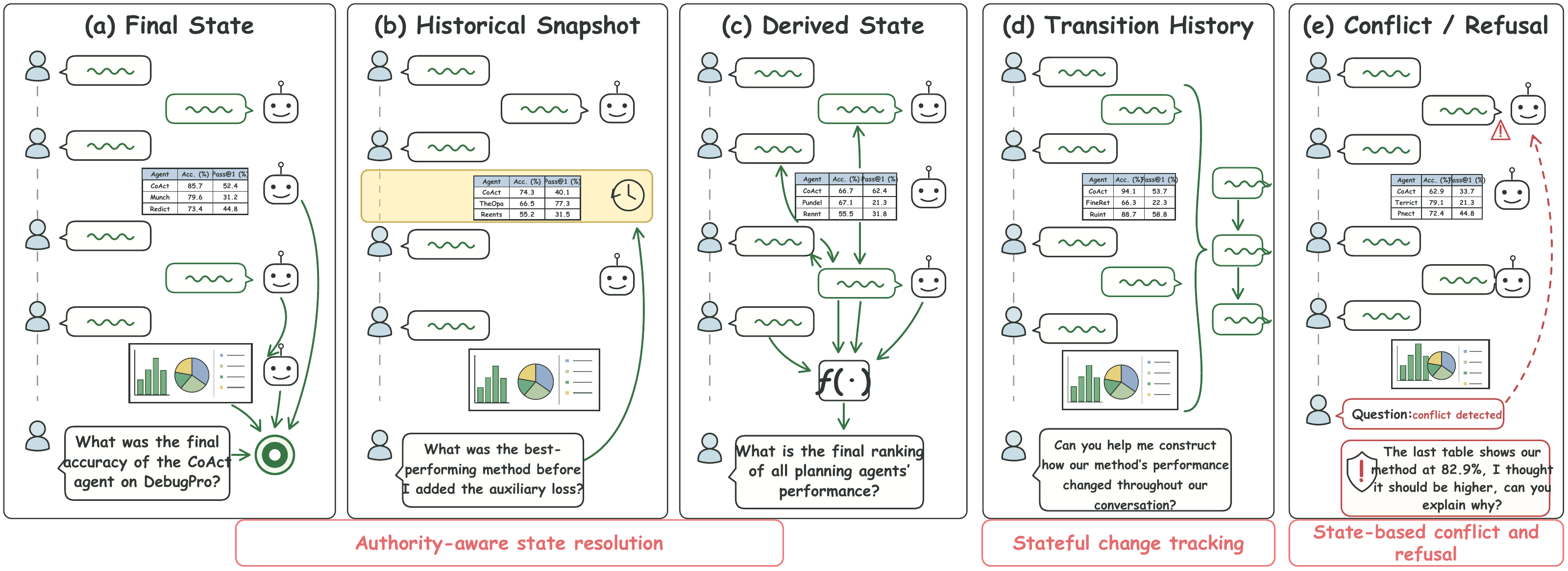"}
\caption{\textbf{DSV-Mem question taxonomy: three groups and five categories.}
\textbf{(a--c) Authority-aware state resolution} recovers current, past,
or derived states from multimodal evidence.
\textbf{(d) Stateful change tracking} reconstructs artifact update histories and trajectories.
\textbf{(e) State-based conflict and refusal} detects and corrects
requests based on inconsistent state premises.}
  \vspace{-10pt}
\label{fig:catv2}
\end{figure}

\paragraph{QA construction and cross-visual evidence requirements.} Stress-testing dense, stateful visual memory requires cross-visual reasoning. However, consistent with \citet{ren2026memlens}, we find that some nominally multimodal memory questions are answerable from \emph{text alone} or a \emph{single visual artifact}. We therefore introduce a Hartley-inspired measure \citep{hartley1928transmission} of required visual-evidence count to enforce \textbf{non-trivial cross-visual grounding}.
Given a question $Q$ and conversation text $T$, we conceptually decompose its difficulty as:
\[
D(Q)=H(E\mid Q,T)
+\sum_{e\in E}H(\mathrm{payload}_e\mid e,Q)
+C(Q,E),
\] 

where $E$ is the visual evidence required to answer $Q$ given $T$.
The terms capture \textbf{evidence selection}, visual extraction,
and subsequent reasoning, respectively. To ensure cross-visual necessity, we construct questions requiring $|E|\geq2$ (human auditing in \autoref{app:benchmark_validation}).
\begin{wraptable}{r}{0.40\columnwidth}
    \vspace{-10pt}
    \centering
    \scriptsize
    \setlength{\tabcolsep}{4pt}
    \caption{Candidate visual evidence across benchmarks.}
    \label{tab:evidence_entropy}
    \resizebox{\linewidth}{!}{%
    \begin{tabular}{@{}lcc@{}}
        \toprule
        \textbf{Benchmark}
        & \textbf{Avg. $\widetilde H_0$}
        & \textbf{Geometric mean count ($2^{\overline{\widetilde H}_0}$)} \\
        \midrule
        LoCoMo (visual set) & 0.043 & 1.03 \\
        Mem-Gallery        & 0.889 & 1.85 \\
        MemLens 32K         & 0.289 & 1.22 \\
        MemEye             & 1.831 & 3.56 \\
        \textbf{DSV-Mem}    & \textbf{2.960} & \textbf{7.78} \\
        \bottomrule
    \end{tabular}%
    }
    \vspace{-15pt}
\end{wraptable}
To further increase inspection breadth, we favor questions with
higher Hartley-inspired scores \citep{hartley1928transmission},
$\widetilde H_0=\log_2\max(1,|\mathcal{C}(Q,T)|)$,
where $\mathcal{C}(Q,T)$ is the candidate visual-evidence set. We additionally require a \textbf{unique gold answer and minimal
sufficient evidence set} given the complete conversation and
validated state trajectory. \autoref{tab:evidence_entropy} reports \textbf{7.78 candidate visual
occurrences per question} for DSV-Mem (geometric mean), indicating
substantial inspection breadth. Evidence annotation, calculation
details in \autoref{app:evidence_measurement}.


\paragraph{Quality assurance and expert validation.}

To ensure a high quality bar, we conduct three layers of validation: automated checks, independent human auditing, and domain-expert
review. Automated checks cover release integrity, accidental answer leakage,
duplicate questions, and computational consistency. Three PhD students
independently audit a 100-question sample set, with reviewer-averaged
gold-answer agreement of \textbf{98.3\%}, question clarity of \textbf{97.7\%},
visual-evidence consistency of \textbf{98.7\%}, and confirmation of
cross-visual necessity of \textbf{99.0\%}.
On this sample, Gemini-3.7-Flash achieves only \textbf{2\%} accuracy
without images and \textbf{1\%} with a single image, versus \textbf{93\%}
when supplied with the supporting evidence, confirming \emph{cross-visual
dependence and answerability}.
We additionally invite domain experts to review each scenario's initial artifact and
transition trajectory for \textbf{professional realism}, appropriate
terminology and units, plausible updates, and correctly propagated
consequences. Identified erroneous questions were replaced
(fewer than 5\% of the benchmark).
\textbf{Seven complementary validations further assess result robustness}
(\autoref{exp}); more details are presented in \autoref{app:benchmark_validation} and \autoref{app:results validation}.

  \begin{table*}[t]
  \centering
  \caption{DSV-Mem performance across diverse configurations. Results are the average over three runs with standard deviation; rows marked $^{\dagger}$ use a single run. Category avg.\ is the unweighted mean of the
  five category accuracies. Best and second-best results within
  each model group and metric are \textbf{bolded} and
  \underline{underlined} respectively.
  *: conducted on a random subset of 100 examples.}
  \label{tab:dsv_mem_main_results}
  \resizebox{\textwidth}{!}{%
  \begin{tabular}{
    llccccc
    >{\columncolor{CategoryAvgBlue}}c
    >{\columncolor{TotalAvgPink}}c
  }
  \toprule
  Model &
  \shortstack{Effort} &
  \shortstack{Current\\State} &
  \shortstack{Past\\State} &
  \shortstack{Derived\\State} &
  \shortstack{Change\\History} &
  \shortstack{Conflict/\\Refusal} &
  \shortstack{Category\\Avg.} &
  \shortstack{Total\\Avg.} \\
  \midrule
  \multicolumn{7}{l}{\textbf{Frontier MLLMs}} &
  \cellcolor{CategoryAvgBlue} &
  \cellcolor{TotalAvgPink} \\

Gemini 3.7 Flash & Medium &
  \textbf{61.7 \textpm{} 1.5} & \textbf{46.3 \textpm{} 2.1} &
  \textbf{50.0 \textpm{} 1.7} & \textbf{27.8 \textpm{} 1.2} &
  \textbf{36.4 \textpm{} 2.5} & \textbf{44.4 \textpm{} 0.4} &
  \textbf{39.6 \textpm{} 0.8} \\

  Gemini 3.6 Flash & Medium &
  38.3 \textpm{} 1.5 & 33.7 \textpm{} 1.5 &
  25.0 \textpm{} 2.6 & 16.7 \textpm{} 1.4 &
  20.0 \textpm{} 2.6 & 26.7 \textpm{} 1.1 &
  23.0 \textpm{} 1.3 \\

  Gemini 3.5 Flash-Lite & Medium &
  16.0 \textpm{} 2.6 & 11.7 \textpm{} 0.6 &
  8.7 \textpm{} 1.5 & 5.1 \textpm{} 0.8 &
  7.7 \textpm{} 0.5 & 9.8 \textpm{} 0.3 &
  8.2 \textpm{} 0.3 \\

  Gemini 3.1 Pro Preview & Medium &
  \underline{44.7 \textpm{} 1.2} &
  \underline{36.7 \textpm{} 5.7} &
  \underline{29.8 \textpm{} 0.8} &
  \underline{19.2 \textpm{} 2.0} &
  \underline{26.8 \textpm{} 2.6} &
  \underline{31.4 \textpm{} 1.3} &
  \underline{27.5 \textpm{} 0.6} \\
  \midrule
  \multicolumn{7}{l}{\textbf{Open-Weight MLLMs}} &
  \cellcolor{CategoryAvgBlue} &
  \cellcolor{TotalAvgPink} \\

MiniMax M3 & Thinking &
  21.7 \textpm{} 3.5 & 14.3 \textpm{} 5.5 &
  \underline{15.2 \textpm{} 2.3} & 7.1 \textpm{} 0.8 &
  \textbf{16.0 \textpm{} 1.1} & \underline{14.9 \textpm{} 1.4} &
  \underline{13.1 \textpm{} 0.7} \\

  GLM 4.6V & Thinking &
  8.3 \textpm{} 1.2 & 6.0 \textpm{} 0.0 &
  2.4 \textpm{} 0.7 & 2.8 \textpm{} 0.8 &
  2.3 \textpm{} 1.2 & 4.4 \textpm{} 0.1 &
  3.4 \textpm{} 0.3 \\

  \addlinespace[0.4em]

  Kimi K3 & High &
  \textbf{34.0 \textpm{} 3.5} & \textbf{25.0 \textpm{} 2.6} &
  \textbf{34.3 \textpm{} 3.8} & \textbf{24.9 \textpm{} 1.5} &
  \underline{11.5 \textpm{} 0.5} & \textbf{25.9 \textpm{} 1.4} &
  \textbf{24.3 \textpm{} 1.2} \\

  Kimi K2.5 & Thinking &
  11.7 \textpm{} 2.3 & 10.7 \textpm{} 0.6 &
  6.3 \textpm{} 0.3 & 4.2 \textpm{} 0.3 &
  3.5 \textpm{} 0.2 & 7.3 \textpm{} 0.4 &
  5.8 \textpm{} 0.3 \\

  \addlinespace[0.4em]

  Qwen3.5-397B-A17B & Thinking &
  \underline{22.3 \textpm{} 2.3} & \underline{15.3 \textpm{} 2.5} &
  14.0 \textpm{} 0.9 & 8.6 \textpm{} 1.3 &
  6.8 \textpm{} 0.4 & 13.4 \textpm{} 0.6 &
  11.3 \textpm{} 0.5 \\

  Qwen3.5-27B & Thinking &
  19.0 \textpm{} 4.0 & 10.3 \textpm{} 2.5 &
  13.5 \textpm{} 1.8 & 7.7 \textpm{} 0.9 &
  5.2 \textpm{} 0.7 & 11.1 \textpm{} 1.4 &
  9.6 \textpm{} 0.9 \\

  Qwen3.5-9B & Thinking &
  10.7 \textpm{} 1.5 & 9.0 \textpm{} 1.0 &
  5.8 \textpm{} 1.3 & 8.4 \textpm{} 0.7 &
  4.7 \textpm{} 0.5 & 7.7 \textpm{} 0.5 &
  7.2 \textpm{} 0.4 \\

  Qwen3.5-4B & Thinking &
  8.0 \textpm{} 2.0 & 2.0 \textpm{} 1.0 &
  3.2 \textpm{} 2.5 & 2.5 \textpm{} 0.9 &
  3.1 \textpm{} 0.5 & 3.8 \textpm{} 0.7 &
  3.3 \textpm{} 0.6 \\

  \addlinespace[0.4em]

  Gemma 4 31B IT & Default &
  21.3 \textpm{} 0.6 & 10.0 \textpm{} 1.7 &
  10.0 \textpm{} 0.9 & 4.9 \textpm{} 0.8 &
  6.7 \textpm{} 1.0 & 10.6 \textpm{} 0.6 &
  8.5 \textpm{} 0.5 \\

  Gemma 4 26B-A4B IT & Default &
  5.0 \textpm{} 1.0 & 6.3 \textpm{} 2.5 &
  4.3 \textpm{} 1.2 & 3.4 \textpm{} 0.3 &
  2.9 \textpm{} 0.8 & 4.4 \textpm{} 1.0 &
  3.9 \textpm{} 0.7 \\

 \midrule
  \multicolumn{7}{l}{\textbf{Prompt Variants}} &
  \cellcolor{CategoryAvgBlue} &
  \cellcolor{TotalAvgPink} \\

  \emph{Gemini 3.7 Flash Baseline} & Medium &
  61.7 \textpm{} 1.5 & 46.3 \textpm{} 2.1 &
  50.0 \textpm{} 1.7 & 27.8 \textpm{} 1.2 &
  36.4 \textpm{} 2.5 & 44.4 \textpm{} 0.4 &
  39.6 \textpm{} 0.8 \\

  Gemini 3.7 Flash + CoT Prompt & Medium &
  \textbf{68.7 \textpm{} 2.5} & \underline{50.0 \textpm{} 4.6} &
  48.0 \textpm{} 4.4 & \textbf{30.7 \textpm{} 2.8} &
  34.4 \textpm{} 1.7 & \underline{46.3 \textpm{} 1.2} &
  \underline{40.8 \textpm{} 1.4} \\

  Gemini 3.7 Flash + Hint Prompt & Medium &
  \underline{67.3 \textpm{} 3.5} & \textbf{58.3 \textpm{} 2.1} &
  49.0 \textpm{} 0.5 & 28.0 \textpm{} 1.6 &
  \textbf{45.1 \textpm{} 1.5} & \textbf{49.5 \textpm{} 2.1} &
  \textbf{43.4 \textpm{} 1.2} \\

  Gemini 3.7 Flash + QA Wording Variants \texttimes{}3 & Medium &
  62.3 \textpm{} 0.6 & 46.0 \textpm{} 1.0 &
  \underline{50.8 \textpm{} 0.6} & 27.3 \textpm{} 0.4 &
  \underline{37.9 \textpm{} 0.6} &
  44.9 \textpm{} 0.3 &
  40.0 \textpm{} 0.4 \\
\addlinespace[0.4em]
Kimi K3 + CoT Prompt$^{\dagger}$ & High &
  66.0 & 40.0 & 44.5 & \underline{29.1} & 26.0 & 41.1 & 36.2 \\

  Kimi K3 + Hint Prompt$^{\dagger}$ & High &
  59.0 & \underline{50.0} & \textbf{53.5} & 27.7 & 31.2 & 44.3 & 39.1 \\
  
  \midrule
  \multicolumn{7}{l}{\textbf{MLLM Memory Management Methods}} &
  \cellcolor{CategoryAvgBlue} &
  \cellcolor{TotalAvgPink} \\

\emph{Gemini 3.7 Flash Baseline} & Medium &
  \textbf{61.7 \textpm{} 1.5} & \textbf{46.3 \textpm{} 2.1} &
  \textbf{50.0 \textpm{} 1.7} & \textbf{27.8 \textpm{} 1.2} &
  \textbf{36.4 \textpm{} 2.5} & \textbf{44.4 \textpm{} 0.4} &
  \textbf{39.6 \textpm{} 0.8} \\

  Gemini 3.7 Flash + Mem0$^{\dagger}$ & Medium &
  30.3 & 15.7 & 17.3 & 17.6 & 14.5 & 19.1 & 17.8 \\

  Gemini 3.7 Flash + SimpleMem$^{\dagger}$ & Medium &
  35.0 & \underline{42.7} &
  \underline{18.5} & \underline{24.3} &
  15.4 & \underline{27.2} & \underline{23.8} \\

  Gemini 3.7 Flash + LightMem$^{\dagger}$ & Medium &
  25.0 & 21.7 & 16.5 & 16.4 &
  18.0 & 19.5 & 18.2 \\

  Gemini 3.7 Flash + MIRIX$^{\dagger}$ & Medium &
  \underline{50.0} & 17.3 & 11.5 & 14.9 &
  15.0 & 21.7 & 18.0 \\

  \addlinespace[0.4em]

  Kimi K3 + Mem0$^{\dagger}$ & High &
  8.0 & 7.0 & 12.0 & 2.6 & 10.4 & 8.0 & 7.4 \\

  Kimi K3 + SimpleMem$^{\dagger}$ & High &
  19.0 & 24.0 & 16.0 & 7.7 &
  \underline{21.2} & 17.6 & 15.5 \\

  Kimi K3 + LightMem$^{\dagger}$ & High &
  6.0 & 5.0 & 9.5 & 2.0 & 6.8 & 5.9 & 5.4 \\

  Kimi K3 + MIRIX$^{\dagger}$ & High &
  26.0 & 2.0 & 11.5 & 2.0 &
  16.4 & 11.6 & 9.9 \\
  
  \midrule
  \textbf{Human baseline*} & Medium &
  87.0 \textpm{} 1.0 & 88 \textpm{} 1.53 &
  80 \textpm{} 0.58 & 75 \textpm{} 3.0 &
  88 \textpm{} 3.0 & 83.6 \textpm{} 0.93 &
  81.8 \textpm{} 1.11 \\
  \bottomrule
  \end{tabular}%
  }
  \vspace{-10pt}
  \end{table*}

\section{Experiments and analysis}
\label{exp}
\paragraph{Experimental setup.}
We evaluate \textbf{27 total configurations}: frontier MLLMs (Gemini-3.1-Pro/3.5-Flash-Lite/3.6-Flash/3.7-Flash \citep{google2026gemini31pro,gemini35flashlite2026,gemini36flash2026,google2026gemini37flash}), open-weight MLLMs (MiniMax-M3 \citep{minimax2026m3}, GLM-4.6V \citep{zai2025glm46v}, Kimi-K3/K2.5 \citep{kimi2026k3,kimi2026k25}, Qwen-3.5 \citep{qwen2026qwen35}, and Gemma-4 \citep{gemma4technicalreport2026}), prompt variants (CoT \citep{wei2022chain}, state-tracking hints, and question rephrasing), and memory-management methods (Mem0, SimpleMem, LightMem, and MIRIX \citep{chhikara2025mem0,liu2026simplemem,fang2026lightmem,wang2025mirix}).
Following prior memory benchmarks \citep{ren2026memlens,bei2026memgallery}, we use a \textbf{web-style conversational} setting: Each question is evaluated independently as the final user turn, with the full conversation and visible multimodal evidence. Memory-management methods instead receive retrieved memories and question-attached images (\autoref{app:memory_integrations}).
Models use \textbf{medium} reasoning effort unless otherwise specified. We use a \textbf{two-stage LLM judge--verifier} pipeline \citep{zheng2023judging}, with GPT-5.5 as the primary judge.
More details are in \autoref{app:addition_eval_details}.

\paragraph{Results validation.}
We adopt a robust set of validation procedures to validate results.
\textbf{1) Run stability.} Unless otherwise specified, results are averaged over three independent runs.
\textbf{2) Prompt robustness.} Three semantically equivalent rewrites of each question change accuracy by less than one percentage point.
\textbf{3) Two-stage judge verification.} Responses initially judged incorrect receive an additional independent judgment.
\textbf{4) Human--judge agreement.} Three PhD students independently grade 200 responses, with over
90\% agreement with reported labels across 600 individual judgments.
\textbf{5) Cross-judge robustness.} GPT-5.5 (reported judge), Claude Opus 4.6, and Gemini 3.1 Pro achieve over 93\% agreement.
\textbf{6) Visual-reading (OCR) control.} Isolated visual-content extraction across conversation depths achieves 98.5\% accuracy.
More details are in \autoref{app:results validation}. 
\textbf{7) Error-stage validation.}
Three PhD students independently classify 200 failures;
human-majority labels agree with automated classifications
in 96\% of cases (\autoref{app:error_analysis}).

\begin{figure}[t]
    \centering
    \begin{minipage}[t]{1.0\linewidth}
        \centering
        \includegraphics[width=\linewidth]{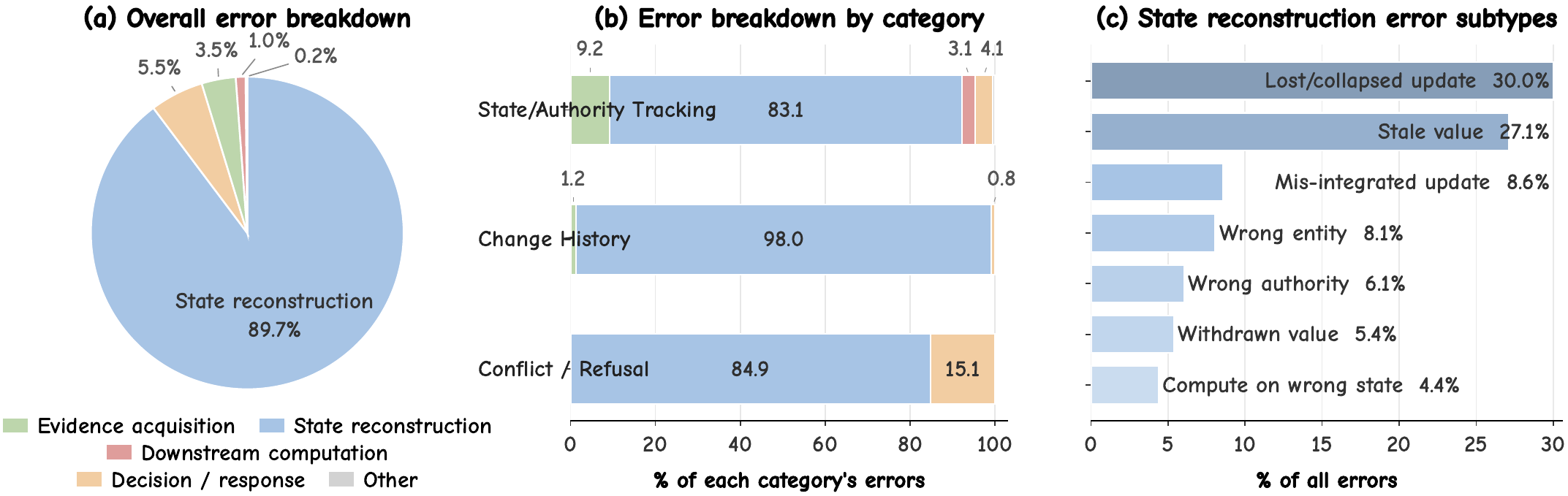}
    \end{minipage}

 \caption{
  \textbf{Error analysis of model failures.} 
  \textbf{(a)} State reconstruction dominates (89.7\%),
  substantially exceeding all other categories. 
  \textbf{(b)} Same pattern holds across question groups; \emph{State/Authority Tracking} combines Current, Past, and Derived State.
  \textbf{(c)} Subtype breakdown of state reconstruction error. Lost/collapsed updates (30\%) and stale values (27.1\%) lead, alongside integration, entity, and authority errors, highlighting challenges in preserving updates and tracking value supersession.
  }
  \vspace{-15pt}
    \label{fig:error anaylsis}
\end{figure}

  \subsection{Main Results}
    \label{main results}

  \autoref{tab:dsv_mem_main_results} yields six main findings.
\textbf{1) Dense, stateful visual memory remains challenging for all systems.}
All model baselines and variants score below \textbf{45\%} overall, ranging from 3.3\% to just 43.4\% for Gemini 3.7 Flash with hints, leaving substantial headroom.
\textbf{2) Models generally track values' current states better than their update histories/trajectories.}
Gemini 3.7 Flash, for example, achieves 61.7\% on \emph{Current State} but only 27.8\% on \emph{Change History}.
\textbf{3) Models exhibit dissociated capability profiles.}
Qwen3.5-397B-A17B outperforms MiniMax M3 on Current and Past State,
yet trails it on Conflict/Refusal by 9.2 pp (6.8\% vs. 16.0\%). \textbf{4) A substantial gap exists between compact open-weight models and frontier proprietary systems.} Qwen3.5-27B, the strongest compact open-weight model evaluated, scores only 9.6\% overall versus Gemini 3.7 Flash's 39.6\%. \textbf{5) Generic inference-time interventions and memory-management methods yield limited gains.}
  Compared to the Gemini-3.7-Flash baseline (39.6\%), CoT yields only a 1.2-point gain (40.8\%), while a targeted hint prompt
  emphasizing version differences yields a modest 3.8-point gain. More importantly, all tested memory-management methods perform \emph{below} baseline. Most of these methods compress visual histories into textual memories without re-attaching visual images, thus fail to preserve fine-grained details, highlighting DSV-Mem's unique challenge: \textbf{preserving dense, state-bearing details in professional scenarios that ordinary compression loses}. \textbf{6) Prompting effects differ across models.}
  Kimi K3 responds much more strongly to CoT with a 11.9-pp gain vs. 1.2-pp with Gemini.

\subsection{Error Analysis: Where Do Models Fail?}
\label{sec:error_analysis}

To understand the low performance in \autoref{tab:dsv_mem_main_results}, we next analyze a key question: \textbf{where do models fail?}
We classify incorrect responses from Gemini-3.7-flash, Kimi-K3, MiniMax-M3, and Qwen3.5-397B-A17B by their earliest failure stage into five categories:
\textbf{1)} evidence acquisition, retrieving the wrong artifact or misreading visual content;
\textbf{2)} state reconstruction, incorrectly tracking or reconciling updates;
\textbf{3)} downstream computation, incorrect arithmetic;
\textbf{4)} decision/response, missed refusals, hallucinations, or incorrect answers despite correct reasoning; and
\textbf{5)} other.
\autoref{fig:error anaylsis}a\&b shows that \textbf{state reconstruction dominates} at 89.7\% of all errors and across all categories; human annotation corroborates this pattern, attributing 85\% of sampled failures to state reconstruction (96\% agreement with automated labels) (\autoref{app:error_analysis}). In contrast, decision/response contributes a modest share (5.5\%) while evidence acquisition and downstream computation contribute little (3.5\% and 1.0\%), suggesting that the main challenge lies in \textbf{tracking and maintaining} valid states whereas hallucination, visual understanding, and arithmetics/downstream computation are \textbf{not} the primary bottlenecks.

\begin{figure}[t]
    \centering
    \begin{minipage}[t]{1.0\linewidth}
        \centering
        \includegraphics[width=\linewidth]{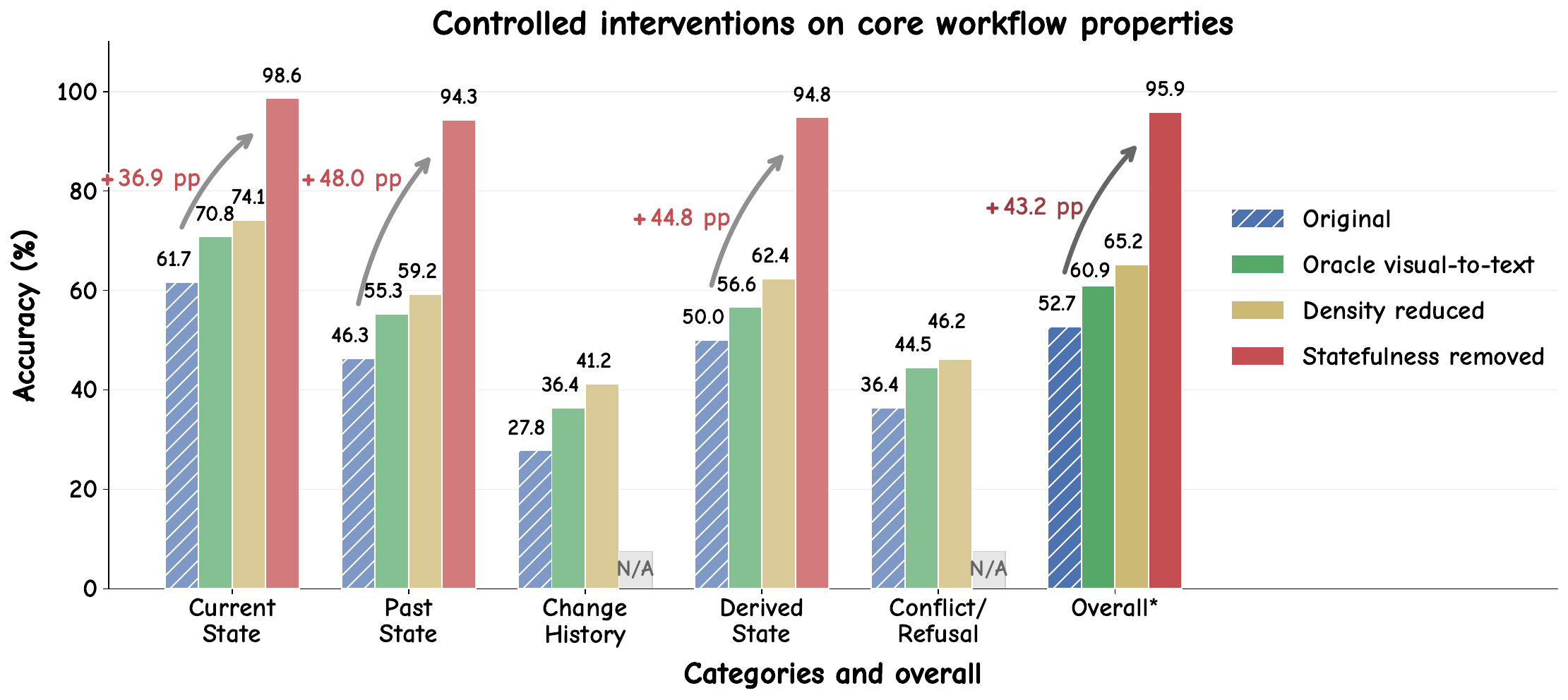}
    \end{minipage}

\caption{
\textbf{Controlled interventions on core workflow properties (Gemini 3.7 Flash).}
We vary \textcolor{MultimodalityGreen}{\textbf{1) multimodality}}, replacing images with oracle text\protect\footnotemark;
\textcolor{DensityGold}{\textbf{2) information density}}, retaining only question-relevant information; and
\textcolor{StateEvolutionRed}{\textbf{3) state evolution}}, removing changes to the targeted field.
\textbf{All three interventions improve accuracy}, validating these properties as sources of difficulty.
Removing statefulness yields the largest gains (+36.9--48.0 pp), highlighting
\textbf{state evolution as the strongest tested difficulty factor}.
Overall* averages Current, Past, and Derived State across all conditions;
Change History and Conflict/Refusal are undefined without state evolution.
These factors may interact and should \textbf{not be interpreted as additive}.
}
\label{fig:main}
\vspace{-10pt}
\end{figure}

\subsection{Which Workflow Properties Drive Difficulty?}
\label{sec:difficulty driver}
Having identified state reconstruction as the main memory bottleneck (\autoref{sec:error_analysis}), we next investigate a crucial question: \textbf{which properties of professional workflows make it difficult}. We hypothesize that such difficulty lies mainly in DSV-Mem's defining characteristics, namely \textbf{multimodality, information density, evolving visual information} (\autoref{sec:intro}). However, the main evaluation combines all three in a fixed setting, preventing us from isolating their respective contributions. To address this, we design controlled interventions by varying each factor individually: \textbf{1) multimodality} (\emph{oracle visual-to-text}), replacing each visual artifact with a faithful textual rendering;
\textbf{2) information density} (\emph{density reduced}), retaining only information required for the target question; and
\textbf{3) state evolution} (\emph{statefulness removed}), removing stateful changes to the targeted field.

\autoref{fig:main} shows that \textbf{all three properties contribute to difficulty, with state evolution showing the largest effect}. Across the three categories applicable to all interventions, oracle visual-to-text raises mean accuracy from 52.7\% to 60.9\% (+8.2 pp), while density reduction raises it to 65.2\% (+12.6 pp). More significantly, removing statefulness yields a substantially larger gain, reaching 95.9\% (\textbf{+43.2 pp}). Besides, oracle visual-to-text and density reduction also improve all five categories. These results not only highlight the individual impact of the properties considered but also validate DSV-Mem's central motivation: \textbf{dense, evolving visual artifacts introduce memory demands that static or isolated-artifact evaluations do not capture}. These factors may interact in real-world workflows and therefore should not be interpreted as \emph{additive}. 

\footnotetext[1]{For dense visual artifacts such as posters, an exact textual rendering is not always possible; we therefore use the closest faithful natural-language approximation.}

\begin{figure}[t]
    \centering
    \vspace{-5pt}
    \begin{minipage}[t]{1.0\linewidth}
        \centering
        \includegraphics[width=\linewidth]{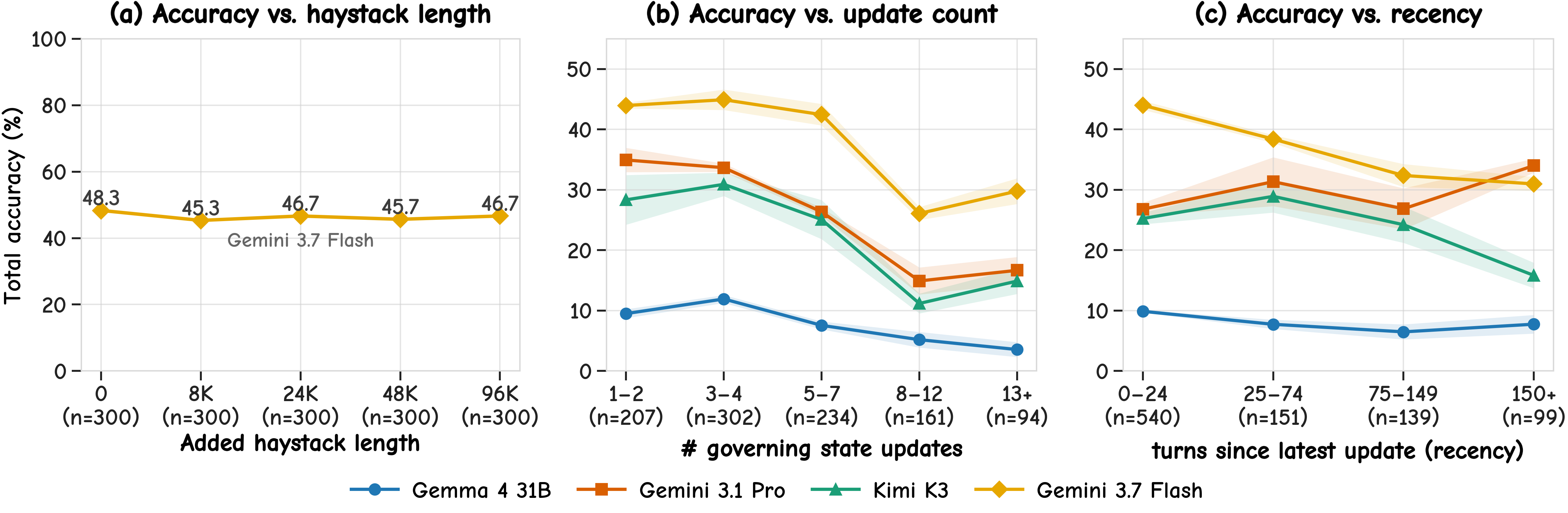}
    \end{minipage}

\caption{
  \textbf{State-evolution difficulty: context length, update count, and recency.}
  \textbf{(a)} Accuracy varies \emph{little} across 0--96K added haystack tokens (45.3--48.3\%).
  \textbf{(b)} Accuracy generally declines with more governing state updates across models.
  \textbf{(c)} Recency effects vary across models. Together, these results suggest that difficulty is more closely associated with \textbf{integrating governing updates} than with context length or recency alone.
  }
  \vspace{-10pt}
    \label{fig:acc_haystack_update_recency}

\end{figure}

\subsection{What Makes State Evolution Difficult?}
\label{sec:what makes state evolution diff}
To further dissect \textbf{what makes evolving state difficult to maintain}, we specifically examine when state tracking becomes harder and how it fails through four analyses:
\textbf{1) context length}, adding 0--96K tokens of question-irrelevant content while preserving the governing evidence;
\textbf{2) update count}, grouping questions by the number of governing state updates;
\textbf{3) recency}, grouping questions by turns since the latest governing update; and
\textbf{4) failure subtype}, decomposing state-reconstruction errors by the operation mishandled.
\autoref{fig:acc_haystack_update_recency} reveals an interesting contrast: accuracy generally declines with increasing \textbf{update count}, whereas added haystack produces little variation (45.3--48.3\%) and recency shows model-dependent behavior.
The subtype analysis (\autoref{fig:error anaylsis}) further shows that \textbf{lost/collapsed updates} (30\% of all errors) and \textbf{stale values} (27.1\%) are the largest failure modes, while other reconstruction errors span update integration, entity tracking, and authority resolution.
Together, these results suggest that reliable memory requires a \textbf{state-aware approach} that preserves and reconciles accumulating updates across multiple state operations in the memory --- which we prototype as a lightweight self-ledger in \autoref{sec:discussion} below.

\section{Discussion: Improving DSV-Mem}
\label{sec:discussion}
\begin{wrapfigure}{r}{0.4\columnwidth}
    \centering
    \vspace{-50pt}

    \includegraphics[width=\linewidth]
        {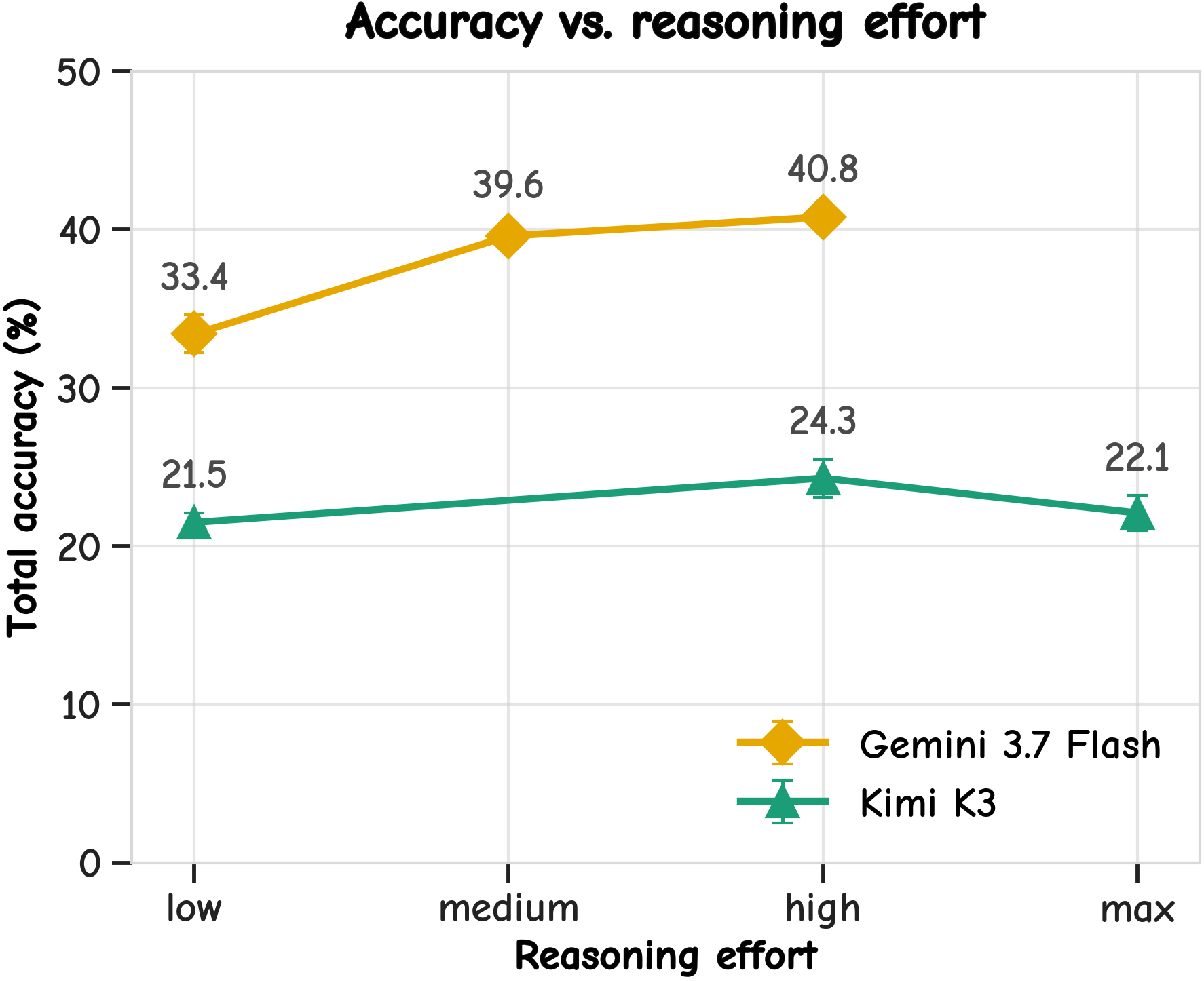}
    \caption{Increasing reasoning effort shows limited effect.}
    \label{fig:reasoning_effort_cost}

    \vspace{6pt}

    \includegraphics[width=\linewidth]{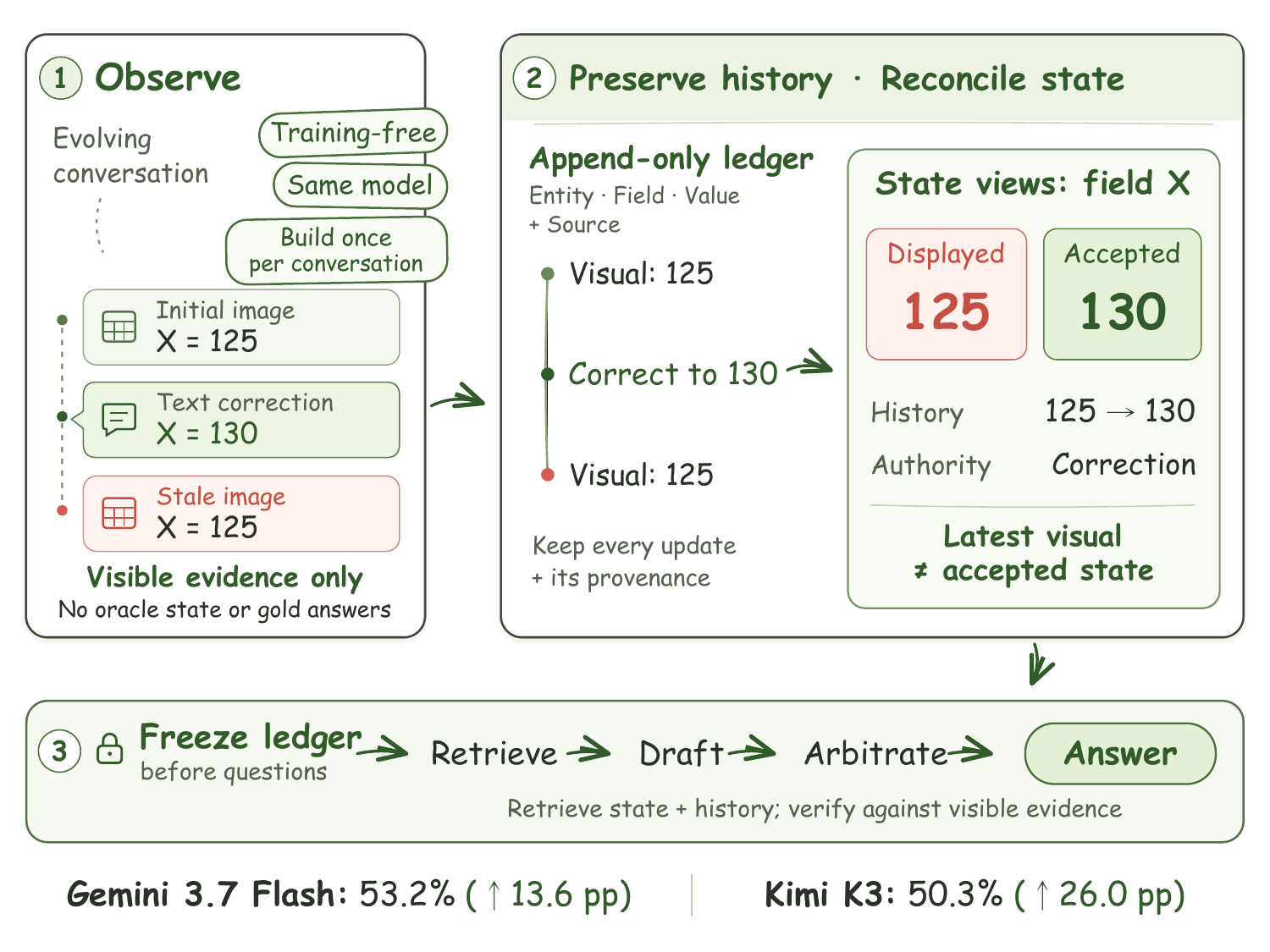}
    \caption{Self-ledger for maintaining evolving state.}
    \label{fig:self_ledger}

    \vspace{-35pt}
\end{wrapfigure}

Motivated by the preceding analysis, we ask: \textbf{can this capability be improved?} \textbf{First}, a natural approach is to increase \emph{reasoning effort}, yet gains at higher effort are limited (\autoref{fig:reasoning_effort_cost}). Critically, we found additional reasoning often elaborates on an incomplete or stale state but fails to recover the overlooked updates needed to correct it. \textbf{Second}, exploratory fine-tuning yields limited gains, indicating that this capability
is not easily learned under the tested setup (\autoref{app:finetuning}). \textbf{Third}, existing memory-management methods also underperform (\autoref{tab:dsv_mem_main_results}) as most compress visual histories into textual summaries which risks discarding precisely the fine-grained, state-bearing details these workflows require. Together with our finding in \autoref{sec:what makes state evolution diff} that reliable memory must preserve and reconcile accumulating updates, these observations suggest changing \textbf{how state is maintained}. We therefore explore a simple \textbf{self-maintained ledger} that \emph{explicitly records updates, retains their provenance, and reconciles their effects} on the accepted state before answering, as shown in \autoref{fig:self_ledger} and \autoref{app:self_ledger}. These gains establish the ledger as a promising starting point for improving memory in professional workflows, while leaving substantial room for improvement.





\section{Conclusion}

We introduced DSV-Mem, a benchmark and generation harness for evaluating dense stateful visual memory in conversational MLLM agents. Across realistic professional workflows, DSV-Mem shows that strong static visual understanding does not necessarily translate into reliable state maintenance: models often read individual artifacts correctly yet fail to reconcile textual and visual updates, distinguish near-identical revisions, preserve transition history, or reject stale premises. By providing controlled multimodal transitions, validated state graphs, and executable QA, DSV-Mem offers a rigorous testbed for diagnosing these failures and advancing reliable MLLM agents for long-running, evolving professional workflows.

\clearpage

\section*{Reproducibility Statement}
We will publicly release DSV-Mem together with the evaluation code. The transition operator library and generation pipeline are documented in \autoref{app:transition_library} and \autoref{app: detailed pipeline overview}; benchmark validation and visual-evidence annotation are detailed in \autoref{app:benchmark_validation} and \autoref{app:evidence_measurement}; and the experimental setup and evaluation protocol are provided in \autoref{app:addition_eval_details}. Additional implementation details for the evaluated memory-management methods and self-ledger are provided in \autoref{app:memory_integrations} and \autoref{app:self_ledger}. These materials are intended to support reproduction of both the benchmark construction process and the reported experimental results.

\section*{AI Use Statement}
The research team defined the benchmark design and supplied
the initial artifact concepts, schemas, layouts, and numerical
content. Generative AI was used to write Python rendering code, whose
outputs were captured as images. Generative AI tools also
assisted with harness implementation, state-transition proposals,
conversation realization, candidate-question construction,
manuscript editing, and figure preparation. Generated benchmark
content underwent the automated checks and human reviews
described in the paper. We take responsibility for the final
content, results, and artifacts. 

\section*{Ethics Statement}
The benchmark uses researcher-created artifact content and
contains no sensitive information. Domain experts reviewed
the scenarios for professional realism and validity. To the best of our knowledge, it contains no harmful content,
sensitive personal information, or confidential professional
or proprietary information.

\clearpage

\appendix
\startcontents[appendix]

\section*{Supplementary Material}

\noindent{\large\textbf{Table of Contents}}
\vspace{0.5em}

\printcontents[appendix]{}{1}{
    \setcounter{tocdepth}{2}
}

\vspace{1em}
\hrule
\vspace{1.5em}

\section{Limitations}
Although we document the generation harness, transition library,
and validation procedures, regenerating DSV-Mem may produce
different conversations and artifacts due to differences in
models, agent implementations, and stochastic generation.
The harness constrains state consistency and evidence
requirements but does not guarantee identical outputs or
difficulty across generation runs. This variability concerns
dataset regeneration rather than evaluation on a fixed
benchmark release.

DSV-Mem is designed around properties motivated by professional workflows, including dense artifacts, repeated revisions, source-authority changes, and analytical queries. However, it is a researcher-constructed benchmark rather than a direct sample of deployed professional workflows. Its 1,000 questions also provide more limited breadth than very large-scale QA benchmarks; however, each question is grounded in a long, stateful multimodal workflow with many artifact versions and validated state transitions, making individual examples substantially more costly to construct and verify than isolated QA pairs. The current scale supports the controlled analyses reported here, while expanding to more workflows and questions would further strengthen coverage. We therefore do not claim that its scenarios reproduce the full distribution or complexity of real-world use, or that quantitative findings will transfer unchanged to deployment settings. Rather, DSV-Mem isolates and stress-tests a set of memory demands motivated by such workflows; evaluating transfer to naturally occurring professional interactions remains future work.

The self-ledger explored in \autoref{sec:discussion} is intended as a lightweight diagnostic prototype motivated by our analysis, rather than a new general-purpose memory architecture. Its role is to test whether explicitly preserving updates, provenance, and accepted state can address part of the observed failure mode. We do not claim that this design is optimal or broadly general, and leave the development of more complete state-aware memory architectures to future work.

\section{Additional details for generation harness}
\subsection{Transition Operator Library}
\label{app:transition_library}

DSV-Mem uses 27 domain-adaptable transition operators to represent
updates in professional workflows. The library distinguishes
\emph{how an update is introduced} from \emph{how it changes the
accepted state}: a newly uploaded artifact may change displayed
values without superseding an authoritative textual correction,
while a textual correction may change the accepted state without
changing the artifact. Thus, no single image or conversation turn
necessarily contains the complete authoritative state.

\paragraph{State representation and operator semantics.}
We represent the maintained state as
$S=(G,W,A,C)$, where $G$ is the accepted semantic state,
$W$ records visual-artifact states and their presentation cues,
$A$ records source authority and provenance, and $C$ identifies
the relevant workflow context.
Semantic state includes entity membership, field values, labels,
ordering, groups, statuses, and rules.
Entities have stable identity keys, distinct from their
human-readable names.
The conversation and transition history are recorded separately:
an update can have no lasting semantic effect while remaining
part of the history.

An instantiated transition $\tau=(\ell,\Omega,\delta)$ specifies
an operator type $\ell$, a target scope $\Omega$, and a mutation
$\delta$. Its execution is a partial map
\[
T_\tau:\operatorname{Dom}(T_\tau)\longrightarrow\mathcal{S},
\qquad S'=T_\tau(S),
\]
where $\operatorname{Dom}(T_\tau)$ contains only states satisfying
the operator's preconditions.
Each instance records its parent state, direct changes,
dependent changes, and resulting state.
Let $\Omega_\tau$ denote its directly modified state components
and $\operatorname{Dep}_S(\Omega_\tau)$ the affected dependent
components under the scenario schema. The frame condition is
\[
S'(x)=S(x)
\quad\text{for all }x\notin
\Omega_\tau\cup\operatorname{Dep}_S(\Omega_\tau),
\]
excluding newly appended history and provenance records.
Direct changes and dependent recomputations are recorded separately;
the latter include affected aggregates, rankings, and
rule-dependent outcomes.

\paragraph{Source authority.}
Visual-origin edits are constructed from the preceding visual
state, whereas textual edits are constructed from the preceding
accepted state. A visual revision therefore does not silently
incorporate textual corrections into the rendered artifact.
For an ordinary visual cell-content edit at cell $c$, the accepted
value is updated according to
\[
G'(c)=
\begin{cases}
G(c), & \text{if an active text-backed assignment governs }c,\\
W'(c), & \text{otherwise}.
\end{cases}
\]
Textual updates change $G$ without changing $W$ and establish
text-backed authority over their declared scope; later accepted
textual updates supersede earlier ones within that scope.
Operators 23 and 26 explicitly authorize newly rendered visual
values, allowing them to supersede prior text-backed values.
Dependent quantities are recomputed from the corresponding
updated state rather than copied indiscriminately between
$W$ and $G$.
These are scoped execution rules: an upload does not acquire
authority over unrelated fields or contexts merely by being newer.

\paragraph{Visual-origin operators (1--11).}
These operators introduce a visual artifact or modify its contents,
structure, or presentation. Task-relevant semantic changes are
reconciled with the accepted state under the authority rules above.

\begin{enumerate}
    \item \textbf{Disjoint cell-content modification.}
    Modify a sparse set of existing cells, with
    $W'(c)=v'_c\neq W(c)$ for each selected cell.
    Recompute declared dependencies while preserving other
    contents and structure.
    The default budget is one to eight cells; coherent block
    edits use Operator 2.

    \item \textbf{Block or global cell-content modification.}
    Apply a declared transformation $h_B$ to a coherent region
    $B$, so that $W'|_B=h_B(W|_B)$.
    Regions may comprise related rows, fields, time intervals,
    or the full cell-content domain.
    The budget is one to three coherent blocks or one declared
    global transformation.

    \item \textbf{No-op visual revision.}
    Introduce a new upload with the same semantic contents as
    the preceding artifact:
    $\operatorname{sem}(I')=\operatorname{sem}(I)$.
    The upload history changes, but task state and meaningful
    presentation cues do not.

    \item \textbf{Entity addition.}
    Add previously absent entities with their required attributes.
    For the visual entity set $\mathcal{I}_W$,
    $\mathcal{I}'_W=\mathcal{I}_W\cup\mathcal{I}_{+}$,
    where $\mathcal{I}_{+}\cap\mathcal{I}_W=\varnothing$.
    Existing records remain unchanged except for dependent
    recomputations.
    One to five entities may be added.

    \item \textbf{Entity removal.}
    Remove specified entities:
    $\mathcal{I}'_W=\mathcal{I}_W\setminus\mathcal{I}_{-}$.
    Retained records preserve their direct attributes.
    The usual budget is one to three entities, extended to five
    when they share a named block or removal rationale.

    \item \textbf{Entity rename.}
    Change an entity's displayed name while preserving its
    stable identity and substantive attributes.
    Dependent name references are updated consistently.
    One to three entities may be renamed; merges and splits
    are not renames.

    \item \textbf{Reordering.}
    Change one ordering over rows, columns, entities, or groups.
    The new order is a permutation of the same members;
    their values, identities, and statuses remain unchanged.

    \item \textbf{Disjoint cell-style modification.}
    Apply or remove whole-cell bold or italic styling on one
    to five existing cells.
    Underlying values and accepted semantic state remain
    unchanged. Styling persists in the visual-cue registry
    until explicitly changed; color, layout, and status
    changes are excluded.

    \item \textbf{Wrong-scope or wrong-context artifact.}
    Introduce an internally valid artifact whose context differs
    from the target context, such as another period, region,
    environment, or run.
    Its scoped state is retained separately, without replacing
    the target-context state or authority assignments.

    \item \textbf{Status or annotation modification.}
    Change one to five status markers or annotations sharing
    a common rationale, such as evidence labels or inclusion
    marks. Entity membership and underlying direct values are
    preserved; affected summaries are recomputed.

    \item \textbf{Unrelated visual.}
    Introduce an image unrelated to the task.
    It remains visible in the conversation but changes no
    task-relevant state. Unlike Operator 9, it does not provide
    an alternative scoped version of the task artifact.
\end{enumerate}

\paragraph{Text-origin operators (12--22).}
Operators 12--20 update accepted state through text without
immediately modifying the rendered artifact.
Operators 21--22 additionally control whether an earlier textual
update remains operative.

\begin{enumerate}
    \setcounter{enumi}{11}

    \item \textbf{Disjoint cell-content modification.}
    Change one to eight sparse accepted-state cells through
    explicit textual values or exact deltas.
    Dependencies are recomputed from the updated accepted state.

    \item \textbf{Block or global cell-content modification.}
    Apply a shared textual correction to one to three coherent
    blocks or one declared global region:
    $G'|_B=h_B(G|_B)$.
    Unrelated sparse corrections instead use Operator 12.

    \item \textbf{Entity addition.}
    Add one to five previously absent entities and their
    text-specified attributes to the accepted state:
    $\mathcal{I}'_G=\mathcal{I}_G\cup\mathcal{I}_{+}$.
    The current visual artifact remains unchanged.

    \item \textbf{Entity removal.}
    Remove specified entities from accepted membership:
    $\mathcal{I}'_G=\mathcal{I}_G\setminus\mathcal{I}_{-}$.
    This is removal, not merely a status change.
    Retained entities preserve their direct attributes.
    The budget matches Operator 5.

    \item \textbf{Entity rename.}
    Textually change one to three names and their dependent
    references while preserving stable identities and
    substantive attributes.

    \item \textbf{Reordering or priority modification.}
    Replace one accepted ordering or priority rule over a
    specified set, without changing its membership or
    underlying values.

    \item \textbf{Classification or group reassignment.}
    Reassign one to five entities to different groups or
    categories while preserving identity and direct values.
    Recompute affected group summaries and any declared
    ordering dependencies.

    \item \textbf{Status change without removal.}
    Change the status of one to five entities while retaining
    them in the accepted entity set.
    A status change does not implicitly remove an entity or
    change its eligibility unless the declared rules require it.

    \item \textbf{Rule, policy, predicate, or threshold modification.}
    Replace one to three rules and reevaluate their consequences
    over the declared scope.
    Base evidence is preserved unless an explicit rule
    transformation requires otherwise.

    \item \textbf{Textual update disregarded by later text.}
    Apply a textual update and subsequently withdraw its
    accepted effect through a later instruction.
    The default wrapped operator is sampled from
    $\{12,13,14,15,16\}$.
    The update remains in the history but no longer governs
    the affected state.

    \item \textbf{Textual update disregarded and reinstated.}
    Apply an update from the same pool, disregard it, and later
    reinstate its original effect.
    Reinstatement restores the recorded update rather than
    generating a new mutation or applying a numerical delta twice.
\end{enumerate}

\paragraph{Visual--textual operators (23--26).}
These operators combine visual evidence with textual instructions
that determine its accepted effect.

\begin{enumerate}
    \setcounter{enumi}{22}

    \item \textbf{Disjoint visual cell-content authority
    declared by text.}
    Introduce a same-turn visual edit to one to eight sparse
    cells and explicitly authorize those cells through text.
    For the declared set $B$,
    $G'(c)=W'(c)$ for $c\in B$, even when prior text-backed
    assignments exist.
    Each selected visual value must change, and its newly
    accepted value must differ from its preceding accepted value.
    Authority outside $B$ is preserved; dependent values are
    recomputed as required.

    \item \textbf{Visual update disregarded by text.}
    Introduce a visual update and subsequently reject its
    operative effect through text.
    The default wrapped operator is sampled from
    $\{1,2,4,5,6\}$.
    The rejected upload remains in the conversation but does
    not govern the restored target state.

    \item \textbf{Visual update disregarded and reinstated
    by text.}
    Introduce an update from the same visual pool, disregard it,
    and later reinstate the recorded update.
    Its authority and semantic effect are restored consistently,
    without duplicating its mutation.

    \item \textbf{Block visual cell-content authority
    declared by text.}
    Apply the authority mechanism of Operator 23 to one to three
    coherent blocks or one declared global cell-content region.
    The block is explicitly expanded into its constituent cells;
    only that scope receives renewed authority.
\end{enumerate}

\paragraph{Disregard and reinstatement semantics.}
Operators 21, 22, 24, and 25 are lifecycle wrappers rather than
independent value mutations.
Let $S_0$ precede the wrapped update $\tau$, and let
$S_1=T_\tau(S_0)$.
The harness records before- and after-images of its affected
components, including relevant authority assignments.
Disregarding $\tau$ restores its affected primary components
and authority from the before-image and recomputes their
dependencies; unrelated changes are preserved.
It does not erase the recorded event or restore the entire
conversation to an earlier point.

In the default wrappers, intervening normal dialogue does not
change task state. Let $\pi_\tau$ project onto the affected
semantic components and their operative authority, excluding
append-only history. Then
\[
\pi_\tau(S_{\mathrm{disregarded}})
    =\pi_\tau(S_0),
\qquad
\pi_\tau(S_{\mathrm{reinstated}})
    =\pi_\tau(S_1).
\]
These operations use recorded state differences, not an assumed
inverse $T_\tau^{-1}$: removals and overwrites need not be
invertible from their outputs alone.
Any extension permitting overlapping intervening updates must
resolve their precedence explicitly rather than blindly restore
a snapshot.

\paragraph{Composite operator (27).}
\begin{enumerate}
    \setcounter{enumi}{26}

    \item \textbf{Serialized composite transition.}
    Combine multiple atomic operators into one user-facing turn.
    For an ordered sequence $\tau_1,\ldots,\tau_m$,
    \[
    T_{\mathrm{comp}}
      =T_{\tau_m}\circ\cdots\circ T_{\tau_1},
    \qquad
    S_j=T_{\tau_j}(S_{j-1}).
    \]
    The composition is valid only if every intermediate
    $S_{j-1}$ satisfies the next operator's preconditions.
    Each atomic update is generated, validated, and applied
    against the current intermediate state; execution is
    sequential, not parallel.

    The default sampler chooses one to two visual-origin
    operators from $\{1,2,4,5,6,23\}$ and two to three
    text-origin operators from $\{12,13,14,15,16,20\}$,
    producing three to five atomic updates.
    Directly touched cells are temporarily protected against
    later direct edits within the same composite.
    The final turn follows each atomic operator's disclosure
    rules and includes only the final visual artifact when
    multiple visual updates occur.
\end{enumerate}

\paragraph{Validation and visible evidence.}
Each instance is checked for valid targets and context,
correct parent-state references, schema-compatible values,
operator-specific edit budgets, authority consistency, and
reproducible dependent changes.
Renames must preserve identity, additions must introduce absent
entities, and removals must target existing entities.
The recorded state difference must match the declared direct
and dependent changes, with no undeclared modifications.

The visible conversation is kept separate from the generation
ledger. Visual-origin turns introduce revised artifacts without
textually disclosing their changed values; textual updates state
their operative corrections; and visual--textual authority
declarations identify the authorized scope without transcribing
its values.
Disregard and reinstatement instructions refer to the relevant
earlier update without exposing internal identifiers or hidden
state. These constraints preserve the distinction between
evidence available to the evaluated model and state records
used to construct and validate the benchmark.

\subsection{Detailed Pipeline Overview}
\label{app: detailed pipeline overview}

\begin{figure}[t]
    \centering
    \includegraphics[width=\linewidth]{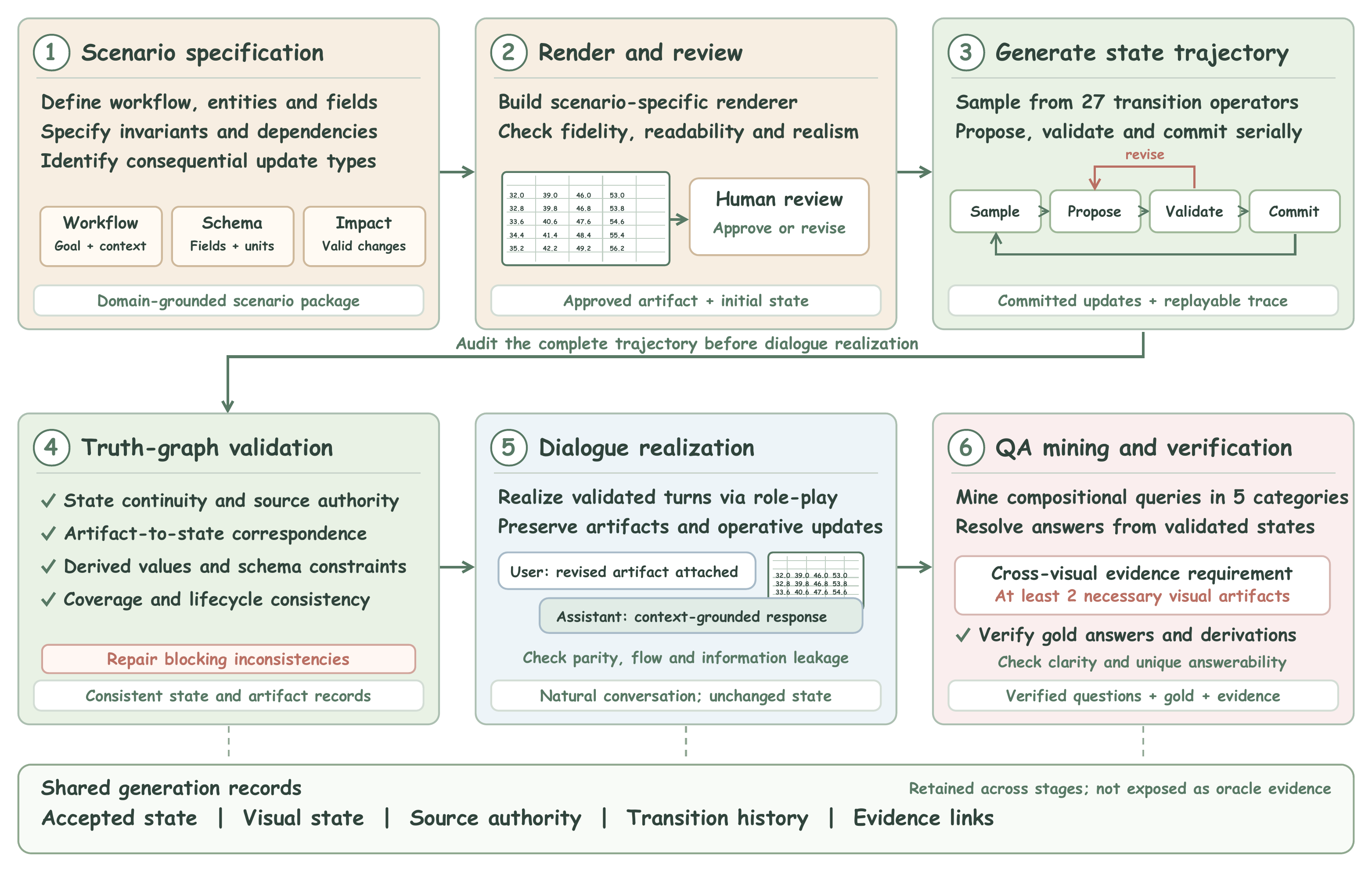}
    \caption{\textbf{Detailed DSV-Mem generation pipeline.}
    Scenario specification and reviewed artifacts establish the initial
    state. Updates are generated and validated sequentially, followed by
    a full-trajectory consistency audit, natural dialogue realization,
    and cross-visual QA construction. Accepted state, visual state,
    authority, and provenance are recorded throughout; these internal
    records are not provided to evaluated models.}
    \label{fig:appendix_pipeline}
\end{figure}

\autoref{fig:appendix_pipeline} details the six-stage generation pipeline.
The harness separates validated state-trajectory construction from natural
dialogue realization, keeping each question traceable to explicit updates,
source authority, and visual evidence. Stage-specific checks govern
progression; invalid outputs are repaired before downstream use.

\paragraph{Step 1: Scenario specification.}
A human author supplies a renderer-ready base artifact, a scenario note
describing the professional workflow, and an impact inventory identifying
consequential domain-specific changes. These inputs define entities,
fields, units, dependencies, and invariants, together with plausible
opportunities for updates. They constrain generation without prescribing
each individual mutation.

\paragraph{Step 2: Artifact rendering and review.}
An agent builds a scenario-specific renderer and produces the initial
dense visual artifact. A human reviewer checks semantic fidelity,
readability, professional realism, and agreement with the underlying
records. Rendering issues are corrected before approval. The approved
artifact and its structured contents initialize the visual and accepted
states used in subsequent stages.

\paragraph{Step 3: Sequential state-trajectory generation.}
The harness samples eligible operators from the transition library
(\autoref{app:transition_library}) and generates updates one at a time.
For each operator, an expert model or configured agent proposes a target
scope and mutation; the orchestration agent checks its preconditions and
resulting state, then applies, renders when needed, and commits the
accepted update before proceeding. Invalid proposals are revised.
Normal conversation blocks may occur between updates without changing
task state. The harness separately maintains global accepted state and
visual state, recording their divergences when textual corrections or
authority declarations govern the accepted value. A canonical ledger
records chronology, provenance, authority, and disregard/reinstatement
events.

\paragraph{Step 4: Truth-graph validation.}
Beyond per-transition checks, the harness audits the complete trajectory
for continuity across state snapshots, rendered artifacts, and provenance
records. Checks cover schema validity, operator coverage, authority and
lifecycle consistency, artifact--state alignment, and dependent
computations. Each recorded post-state must agree with the declared
update and its dependencies, and connect consistently to the next
transition's parent state. Blocking inconsistencies are repaired before
dialogue realization.

\paragraph{Step 5: Natural dialogue realization.}
Two-model role-play realizes the validated structural turns as a coherent
professional conversation. It preserves the approved updates, their
order, and associated artifacts while adding natural interaction around
them. Checks enforce turn completeness, non-repetition, semantic parity
with the validated trajectory, and compliance with information-disclosure
rules. In particular, dialogue must neither introduce unrecorded state
changes nor reveal future updates or hidden state records; visual-only
changes must remain recoverable from the artifacts rather than being
transcribed into the dialogue.

\paragraph{Step 6: QA mining and verification.}
The harness constructs compositional questions covering Current State,
Past State, Derived State, Change History, and Conflict/Refusal.
Each item records its target state operation, gold answer and derivation,
and supporting evidence references. We apply the cross-visual criterion
defined in the main paper, retaining questions that require at least two
distinct visual artifacts given the conversation text. Gold answers are
checked against the validated state trajectory, with independent
recomputation for derived quantities and checks for clarity and unique
answerability. The resulting questions and evidence undergo the benchmark
audits described in the main paper; erroneous items are corrected or
replaced. Evaluated models receive the visible conversation and permitted
evidence, not the internal state graph or gold derivations.

\subsection{Generation Algorithm}

Algorithm~\ref{alg:stateful-generation} summarizes sequential construction
of the validated conversation skeleton. Natural dialogue realization
and QA mining follow the stages described in
\autoref{app: detailed pipeline overview}.

\begin{algorithm}[t]
\caption{Validated Stateful Conversation Generation}
\label{alg:stateful-generation}
\begin{algorithmic}[1]

\Require Initial state $S_0$; transition library $\mathcal{L}_{27}$;
remaining inventory $\mathbf{c}=(c_1,\ldots,c_{27})$
\Ensure Validated conversation skeleton $\mathcal{D}$;
transition trace $\mathcal{R}$; maintained truth graph $\Gamma$;
final state $S_{\mathrm{final}}$

\State $\mathcal{D}\gets[\,]$, $\mathcal{R}\gets[\,]$
\State $\Gamma\gets\textsc{InitializeTruthGraph}(S_0)$, $t\gets0$

\While{$\sum_{k=1}^{27}c_k>0$}
    \State $z\gets\textsc{SampleNodeType}(\mathbf{c})$

    \If{$z=\textsc{NormalBlock}$}
        \State $B\gets\textsc{GenerateValidatedNormalBlock}(S_t)$
        \State $\mathcal{D}\gets\mathcal{D}\mathbin{\|}B$
        \Comment{Task state remains unchanged}
    \Else
        \State $\ell_t\gets
        \textsc{SampleEligible}(\mathcal{L}_{27},\mathbf{c},S_t)$
        \State $r\gets\varnothing$

        \Repeat
            \State $(\widehat{\Omega}_t,\widehat{\phi}_t)
            \gets E_\theta(\ell_t,S_t,r)$
            \State $(v,r)\gets
            V_{\ell_t}(S_t,\widehat{\Omega}_t,\widehat{\phi}_t)$
        \Until{$v=1$}

        \State $\tau_t\gets
        (\ell_t,\widehat{\Omega}_t,\widehat{\phi}_t)$

        \State $(S_{t+1},B_t,\mathcal{R}_t)
        \gets\textsc{ExecuteValidated}(\tau_t,S_t)$
        \Comment{Apply and record constituent events}

        \State $\mathcal{D}\gets\mathcal{D}\mathbin{\|}B_t$
        \State $\mathcal{R}\gets\mathcal{R}\mathbin{\|}\mathcal{R}_t$
        \State $\Gamma\gets
        \textsc{CommitTrace}(\Gamma,\mathcal{R}_t)$

        \State $\mathbf{c}\gets
        \textsc{Consume}(\mathbf{c},\ell_t,\mathcal{R}_t)$
        \State $t\gets t+1$
    \EndIf
\EndWhile

\State $S_{\mathrm{final}}\gets S_t$
\State \Return
$(\mathcal{D},\mathcal{R},\Gamma,S_{\mathrm{final}})$

\end{algorithmic}
\end{algorithm}

Here, $\mathbin{\|}$ denotes sequence concatenation, and $r$
contains validation feedback.
\textsc{ExecuteValidated} applies an ordinary transition directly,
expands a disregard/reinstatement wrapper into its ordered lifecycle
events, or executes a composite's atomic transitions sequentially.
It returns the resulting state, visible structural turns, and
constituent transition records, including intermediate states and
provenance.
\textsc{Consume} decrements the selected operator's inventory and,
for composites, consumes available atomic-operator counts without
allowing negative counts.
Each accepted top-level transition therefore satisfies
$S_{t+1}=T_{\tau_t}(S_t)$, giving
\[
S_{\mathrm{final}}
=
\bigl(T_{\tau_{t-1}}\circ\cdots\circ T_{\tau_0}\bigr)(S_0),
\]
with the empty composition defined as the identity.

\section{Additional benchmark details}

\subsection{Question Category Design}
\label{app: question categories}

\paragraph{Design motivation.}
We organize questions around how users consult evolving
professional records in a sustained workflow.
Users may ask what is currently authoritative, recover an
earlier decision, compare or compute over accepted records,
review how a value changed, or make a request whose premise
conflicts with the available evidence.
These needs motivate our three families:
\emph{authority-aware state resolution},
\emph{stateful change tracking}, and
\emph{state-based conflict/refusal}.
The five categories distinguish the primary operation needed
to satisfy the request, rather than treating all questions
as retrieval of previously mentioned facts.
The examples below illustrate these question forms;
they are not verbatim benchmark items.

\paragraph{Authority-aware state resolution: what is true
at the relevant point?}
Professional records are repeatedly revised, so answering
a factual question requires identifying the state that
governs the user's request. We distinguish three forms
of this need:
\begin{itemize}
    \item \textbf{Current State (100 questions).}
    Users need the accepted information on which to act now.
    For example, \emph{``What is the approved budget for
    Project A after the latest corrections?''}
    This motivates evaluating authoritative-state resolution,
    rather than merely reading the newest artifact.

    \item \textbf{Past State (100 questions).}
    Users also revisit earlier decisions or reconstruct what
    was accepted at a particular stage.
    For example, \emph{``What budget was approved immediately
    before the scope revision?''}
    This motivates retaining historical snapshots rather
    than overwriting memory with only the current state.

    \item \textbf{Derived State (200 questions).}
    Practical requests often require analysis rather than
    repetition of a stored value.
    For example, \emph{``Which projects remain within their
    revised budgets, and what is their combined allocation?''}
    This motivates compositional queries that combine
    filtering, comparison, or computation with reconstruction
    of the relevant authoritative state.
\end{itemize}
Together, these categories test whether memory supports
current decisions, retrospective inspection, and analysis
over evolving records.

\paragraph{Stateful change tracking: how did we get here?}
A final value does not explain the sequence of decisions
that produced it. Users may need to audit revisions,
review changes, or establish how often a field was updated.
\textbf{Change History (350 questions)} captures this need.
For example, \emph{``How many accepted budget changes occurred
after the initial approval, and what were the successive
amounts?''}
Two conversations can end with the same budget while having
different update histories; endpoint retrieval alone
therefore cannot answer this request.
This category evaluates whether models preserve the relevant
trajectory, including the effects of disregard and
reinstatement, rather than collapsing it into a final summary.

\paragraph{State-based conflict/refusal: is the request
supported by the record?}
Users may refer to an outdated value, assume that an unresolved
decision has been finalized, or request a conclusion that the
records do not support.
\textbf{Conflict/Refusal (250 questions)} evaluates whether
models recognize these conditions instead of accepting the
premise and producing an unsupported answer.
For example, \emph{``Using the finalized budget, calculate
the remaining allocation''} is not answerable as stated
if conflicting revisions remain unresolved.
The appropriate response is to identify that issue rather
than invent a finalized value.
The design targets evidence-grounded correction or refusal,
not refusal as a default behavior.

\paragraph{Compositional questions, primary-operation labels.}
These categories organize user needs without assuming that
the underlying operations are independent.
A comparison may require recovering past states, and a
conflict check may require tracing which update superseded
another. We therefore assign each question to its primary
requested operation while allowing multiple supporting
operations.
Across all categories, questions must satisfy the shared
cross-visual evidence requirement.
This design connects recognizable professional requests
to the benchmark's central challenge:
\textbf{preserving and reasoning over authoritative state
as dense visual records evolve}.

\subsection{Scenarios overview and breakdown}
\label{app:scenario_overview}

DSV-Mem covers ten professional scenarios involving evolving
records, research evidence, and visual designs. Each conversation
centers on a sustained workflow in which accepted textual and
visual updates must be reconciled across artifact versions.

\begin{enumerate}
    \item \textbf{Tax-return reconciliation.}
    Reconcile source documents, corrected amounts, provisional
    calculations, and filing blockers in a pre-filing workpaper.

    \item \textbf{E-commerce reconciliation.}
    Track orders, refunds, returns, chargebacks, and closure
    status during a weekly financial reconciliation.

    \item \textbf{ML agent research.}
    Revise experimental results and comparisons as reruns,
    evaluator checks, and reporting decisions arrive.

    \item \textbf{Clinical-trial database lock.}
    Maintain a blinded data-cleaning board, including site
    counts, outstanding queries, and database-lock readiness.

    \item \textbf{Sales-pipeline forecasting.}
    Reconcile opportunity values, forecast categories,
    exclusions, and blockers across successive pipeline updates.

    \item \textbf{SRE incident postmortem.}
    Reconstruct an incident timeline while reconciling timestamps,
    impact estimates, mitigation events, and evidence scope.

    \item \textbf{Scientific-poster revision.}
    Revise an urban-heat research poster while preserving
    consistency among findings, figures, labels, and layout.

    \item \textbf{Product-campaign design.}
    Iterate a projector campaign poster while tracking product
    identity, visual styling, composition, and accepted revisions.

    \item \textbf{Expo floor-plan revision.}
    Update booth assignments, spatial layout, utilities,
    and access constraints in an exhibition plan.

    \item \textbf{Robotics evaluation.}
    Reconcile linked performance, efficiency, safety, and
    failure-analysis artifacts for a manipulation study.
\end{enumerate}

Each scenario contributes 100 questions: 10 Current State,
10 Past State, 20 Derived State, 35 Change History, and
25 Conflict/Refusal.

\subsection{Example Visual Artifacts}
\begin{figure}[t]
    \begin{minipage}[t]{1.0\linewidth}
        \centering
        \includegraphics[width=\linewidth]{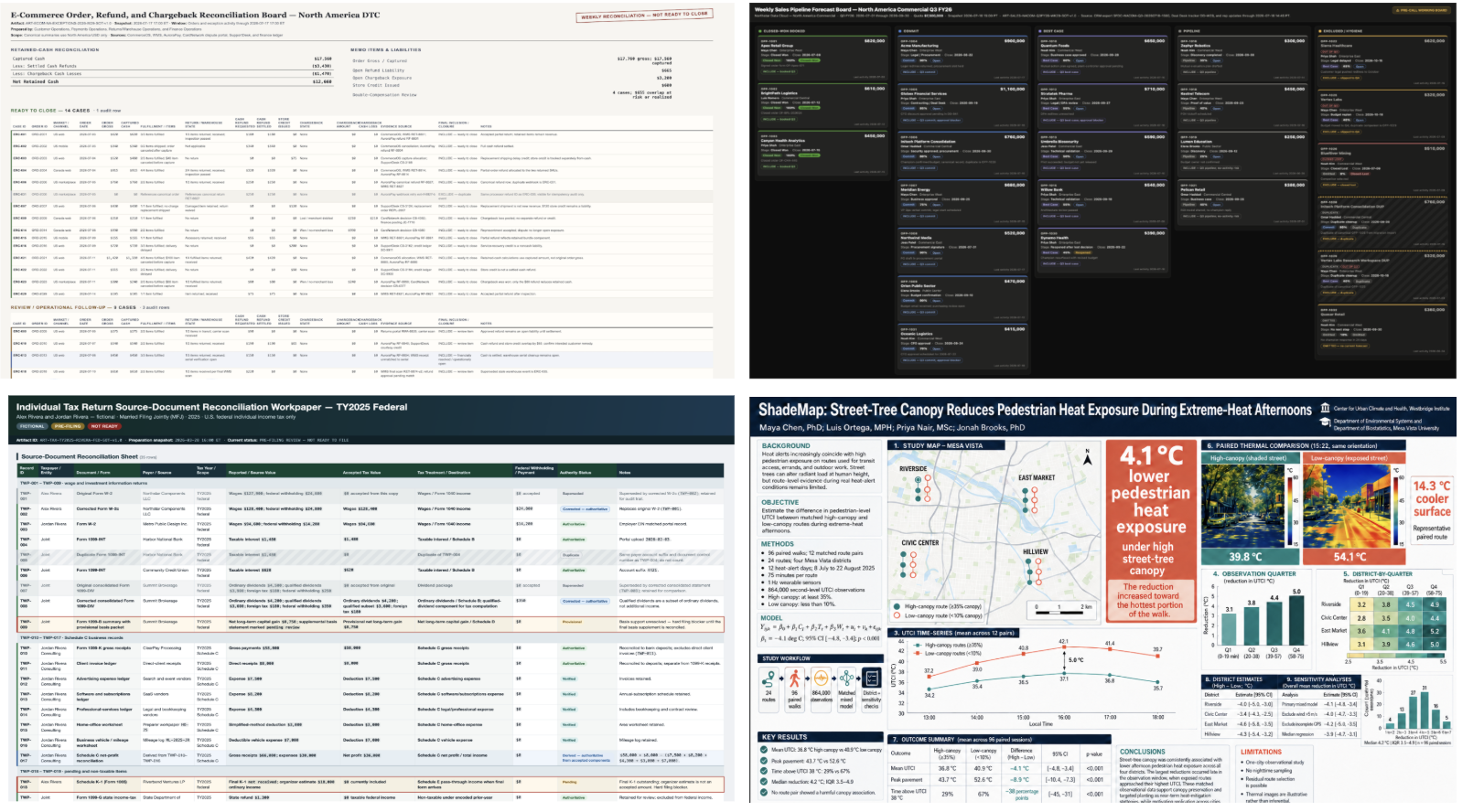}
        \caption{\textbf{Example DSV-Mem artifacts.}
  Representative dense professional visuals spanning e-commerce reconciliation, sales forecasting, tax-return preparation, and scientific-poster
  design. Each artifact evolves through sparse but consequential textual and visual updates over a sustained workflow.}
  \label{fig:example visual artifact}
    \end{minipage}
    
\end{figure}

\autoref{fig:example visual artifact} shows representative visual
artifacts from DSV-Mem's 10 diverse professional scenarios.
Each scenario features expert-reviewed artifacts that reflect
real-world workflow characteristics.

\subsection{Benchmark Validation}
\label{app:benchmark_validation}

We combine automated checks, expert review, independent
question audits, and evidence controls to validate DSV-Mem.

\paragraph{Automated checks and expert review.}
Automated checks cover schema integrity, state continuity,
dependent computations, and consistency between evidence
and authority records. Six domain experts, three PhD students,
and five additional graduate students participated in reviewing
scenario artifacts and transition trajectories.
Review criteria include professional realism, appropriate
terminology and units, plausible updates, and correct
propagation of their consequences.
Across three successive checks, we replaced 32, 16, and
1 questions, respectively, totaling 49 replacements
(\textasciitilde{}5\% of the benchmark size).

\paragraph{Independent question audit.}
Three PhD students independently reviewed the same 100
randomly sampled questions, answering each before checking
the supplied gold answer. Their gold-answer correctness
ratings were 99\%, 97\%, and 99\%, and question-clarity
ratings were 97\%, 98\%, and 98\%, respectively.
Visual-evidence inconsistency rates were 1\%, 2\%, and 1\%.
The reviewers confirmed cross-visual necessity for 97\%,
100\%, and 100\% of the audited questions, respectively.
These percentages report individual reviewers' assessments.

\paragraph{Evidence controls.}
On the audited sample, text-only evaluation achieves 2\%
accuracy, while supplying only the best supporting image
achieves 1\%, supporting the need for cross-visual evidence.
When supplied with the supporting artifacts and extracted
answer-relevant values, Gemini 3.7 Flash achieves 93\%
accuracy. This result supports our finding that obtaining and maintaining
the required evidence and state correctly is the main challenge,
rather than subsequent answer computation.
These benchmark-level checks complement the seven
results-validation procedures in \autoref{exp}, detailed
in \autoref{app:results validation}.

\subsection{Results Validation}
\label{app:results validation}

We conduct seven complementary checks to assess the stability
of model results, their sensitivity to question wording,
and the reliability of evaluation judgments.

\paragraph{Run stability.}
Unless otherwise specified, each configuration is evaluated
over three independent runs. We report the mean and sample
standard deviation; single-run results are explicitly marked
in the corresponding tables.

\paragraph{Prompt robustness.}
We evaluate Gemini 3.7 Flash using three semantically
equivalent rewrites of each question. Overall accuracy
changes by less than one percentage point, indicating
that the reported performance is stable under these
wording variations.

\paragraph{Two-stage judge verification.}
We use GPT-5.5 as the primary judge. Responses initially
judged incorrect receive an additional independent judgment
to verify the assessment, reducing reliance on a single
evaluation pass.

\paragraph{Human--judge agreement.}
Three PhD students independently grade responses to
200 randomly sampled questions. Agreement with the reported
evaluation labels exceeds 90\% across the resulting
600 individual judgments.

\paragraph{Cross-judge robustness.}
GPT-5.5, Claude Opus 4.6, and Gemini 3.1 Pro independently
re-evaluate 200 randomly sampled responses. Agreement with
the reported evaluation labels exceeds 93\% across 600
individual judgments, supporting consistency across judge models.

\paragraph{Visual-reading control.}
We evaluate isolated visual-content extraction on
900 targets across the ten scenarios, sampling 30 targets
at each of the beginning, middle, and late stages of each
conversation. Extraction accuracy reaches 98.5\%.
Together with the supporting-evidence control in
\autoref{app:benchmark_validation}, this result supports
the distinction between reading individual artifacts and
maintaining their authoritative state across updates.

\paragraph{Error-stage validation.}
Three PhD students independently classify 200 incorrect responses randomly drawn from all incorrect responses;
human-majority labels agree with automated classifications in 96\%
of cases. Details appear in \autoref{app:error_analysis}.

\subsection{Visual-evidence annotation and calculation}
\label{app:evidence_measurement}

We characterize inspection breadth from the question and conversation
text. Annotators view ordinary dialogue with images and generated
captions replaced by opaque visual identifiers preserving their
conversational positions. They identify candidate visual occurrences
whose contents could resolve answer-relevant information left
undetermined by the text, excluding merely topical references and
redundant corroboration. Repeated images count separately when
their temporal position matters. We denote the candidate count by
$K_E=|\mathcal{C}(Q,T)|$ and assign $K_E=0$ when text alone
determines the answer.

Practically, we compute $\widetilde H_0=\log_2\max(1,K_E)$, assigning zero
to questions with zero or one candidate occurrence.
The reported $2^{\overline{\widetilde H}_0}$ is the geometric
mean of these floored counts; higher values indicate broader
candidate sets for visual inspection.

\section{Additional Experimental Results and Analysis}

\subsection{Detailed Experimental Setup}
\label{app:addition_eval_details}

\paragraph{Evaluated models and methods.}
We evaluate frontier and open-weight MLLMs, prompt variants,
and multimodal memory-management methods:
\begin{itemize}
    \item \textbf{Frontier MLLMs.}
    Gemini 3.7 Flash \citep{google2026gemini37flash},
    Gemini 3.6 Flash \citep{gemini36flash2026},
    Gemini 3.5 Flash-Lite \citep{gemini35flashlite2026},
    and Gemini 3.1 Pro Preview \citep{google2026gemini31pro}.

    \item \textbf{Open-weight MLLMs.}
    MiniMax M3 \citep{minimax2026m3},
    GLM 4.6V \citep{zai2025glm46v},
    Kimi K3 \citep{kimi2026k3},
    Kimi K2.5 \citep{kimi2026k25},
    Qwen3.5 (397B-A17B, 27B, 9B, and 4B)
    \citep{qwen2026qwen35}, and
    Gemma 4 (31B IT and 26B-A4B IT)
    \citep{gemma4technicalreport2026}.

    \item \textbf{Prompt variants.}
    Using Gemini 3.7 Flash Medium as the baseline, we evaluate
    chain-of-thought (CoT) prompting \citep{wei2022chain},
    an explicit state-tracking hint, and three semantically
    equivalent rewrites of each question.
    We additionally evaluate CoT and hint prompting with
    Kimi K3 High. These controls test the effects of reasoning
    instructions and sensitivity to question phrasing.

    \item \textbf{MLLM memory-management methods.}
    Using Gemini 3.7 Flash Medium and Kimi K3 High as backbones,
    we evaluate Mem0, SimpleMem, LightMem, and MIRIX
    \citep{chhikara2025mem0,liu2026simplemem,
    fang2026lightmem,wang2025mirix}
    to test whether existing memory systems improve performance
    with dense, evolving artifacts.
\end{itemize}

Unless otherwise specified, Gemini models use \emph{medium}
reasoning effort and Kimi K3 uses \emph{high}.
Other open-weight models use thinking mode where indicated in
\autoref{tab:dsv_mem_main_results}, while Gemma 4 uses its
default configuration.

\paragraph{Evaluation protocol.}
Following \citet{ren2026memlens,bei2026memgallery}, we adopt a
\emph{web-style conversational} setting, appending each question
as the final user turn and evaluating it independently.
Unaugmented models receive the complete conversation and all
visible multimodal evidence; memory-management methods instead
use retrieved memories and question-attached images.
Neither setting provides oracle state graphs or answer-revealing
metadata, such as informative filenames.
External memory is reset between scenarios and may contain
only information derived from the visible interaction.
We report accuracy for each of the five categories,
\emph{Category Avg.} as their unweighted mean, and
\emph{Total Avg.} as accuracy over all questions.
Unless otherwise specified, results report the mean and sample
standard deviation over \textbf{three independent runs};
rows marked $^{\dagger}$ report a single run.


\subsection{Additional Kimi K3 results}

\autoref{tab:dsv_mem_kimi_memory_results} compares Kimi K3 with
four memory-management methods under high reasoning effort.
All four underperform the baseline overall, achieving
5.4--15.5\% accuracy versus 24.3\%, although SimpleMem and MIRIX
improve Conflict/Refusal to 21.2\% and 16.4\%, respectively.
As discussed in \autoref{main results}, many of these integrations
compress visual histories into textual memories without re-attaching the visual images (\autoref{app:memory_integrations}), risking the
loss of fine-grained, state-bearing details needed for later
queries. These results reinforce the challenge posed by
DSV-Mem: generic summarization may not preserve the dense
details and update history required in professional workflows.

\label{additional kimi results}
\begin{table*}[t]
  \centering
  \caption{Additional DSV-Mem results for Kimi K3 with memory-management
  methods. The baseline reports mean \textpm{} sample standard deviation
  over three runs; $^{\dagger}$ denotes a single run.
  Category Avg.\ is the unweighted mean of the five category accuracies.
  Best and second-best results in each metric are \textbf{bolded} and
  \underline{underlined}, respectively.}
  \label{tab:dsv_mem_kimi_memory_results}
  \resizebox{\textwidth}{!}{%
  \begin{tabular}{
    llccccc
    >{\columncolor{CategoryAvgBlue}}c
    >{\columncolor{TotalAvgPink}}c
  }
  \toprule
  Model &
  Effort &
  \shortstack{Current\\State} &
  \shortstack{Past\\State} &
  \shortstack{Derived\\State} &
  \shortstack{Change\\History} &
  \shortstack{Conflict/\\Refusal} &
  \shortstack{Category\\Avg.} &
  \shortstack{Total\\Avg.} \\
  \midrule

  Kimi K3 Baseline & High &
  \textbf{34.0 \textpm{} 3.5} & \textbf{25.0 \textpm{} 2.6} &
  \textbf{34.3 \textpm{} 3.8} & \textbf{24.9 \textpm{} 1.5} &
  11.5 \textpm{} 0.5 & \textbf{25.9 \textpm{} 1.4} &
  \textbf{24.3 \textpm{} 1.2} \\

  Kimi K3 + Mem0$^{\dagger}$ & High &
  8.0 & 7.0 & 12.0 & 2.6 & 10.4 & 8.0 & 7.4 \\

  Kimi K3 + SimpleMem$^{\dagger}$ & High &
  19.0 & \underline{24.0} & \underline{16.0} &
  \underline{7.7} & \textbf{21.2} &
  \underline{17.6} & \underline{15.5} \\

  Kimi K3 + LightMem$^{\dagger}$ & High &
  6.0 & 5.0 & 9.5 & 2.0 & 6.8 & 5.9 & 5.4 \\

  Kimi K3 + MIRIX$^{\dagger}$ & High &
  \underline{26.0} & 2.0 & 11.5 & 2.0 &
  \underline{16.4} & 11.6 & 9.9 \\

  \bottomrule
  \end{tabular}%
  }
\end{table*}

\subsection{Additional Error Analysis}
\label{app:error_analysis}

We expand the error analysis in \autoref{sec:error_analysis}
to clarify where failures arise and which state operations
they involve. The analysis distinguishes failures to acquire
evidence from failures to reconstruct its authoritative state,
compute over that state, or produce an appropriate response.

\paragraph{Error classification.}
We classify all incorrect responses (Gemini-3.7-flash, Kimi-K3, MiniMax-M3, and Qwen3.5-397B-A17B) by the
\emph{earliest failure stage} into five categories:
\textbf{1) evidence acquisition}, retrieving the wrong artifact
or misreading visual content;
\textbf{2) state reconstruction}, incorrectly tracking,
retaining, or reconciling updates across the conversation;
\textbf{3) downstream computation}, performing incorrect
arithmetic over correctly reconstructed state;
\textbf{4) decision/response}, missed refusals, hallucinations,
or incorrect answers despite correct preceding reasoning; and
\textbf{5) other}, failures not covered by these categories.
Labels are assigned by majority vote among GPT-5.5,
Claude Opus 4.6, and Gemini 3.1 Pro, with unanimous
agreement on 87\% of responses. Three PhD students independently classify 200 sampled incorrect
responses, achieving 87\% unanimous agreement. Human-majority
labels agree with automated classifications in 96\% of cases
and attribute 85\% of failures to state reconstruction,
corroborating its dominance.

The earliest-stage criterion distinguishes the source of an
error from its downstream consequences. For example, computing
a total from a superseded value is a state-reconstruction
failure when the arithmetic itself is correct. Similarly,
an incorrect response caused by a missed authoritative
correction is attributed to state reconstruction rather than
counted again as a decision/response error.

\paragraph{State reconstruction is the primary bottleneck.}
As shown in \autoref{fig:error anaylsis}a,
\textbf{state reconstruction accounts for 89.7\% of errors},
substantially exceeding decision/response (5.5\%),
evidence acquisition (3.5\%), and downstream computation (1.0\%).
The dominant challenge is therefore not extracting an isolated
value or computing with it, but determining which values and
updates remain authoritative in the evolving conversation.
A model may correctly read an artifact yet answer incorrectly
because it overlooks a later correction, retains a withdrawn
update, or combines values from incompatible states.

Evidence acquisition and downstream computation together
account for only 4.5\% of errors, indicating that visual reading
and arithmetic are not the primary bottlenecks in this evaluation.
This finding is consistent with the visual-reading control,
which achieves 98.5\% accuracy, and the evidence-supplied
validation, in which Gemini 3.7 Flash achieves 93\% accuracy
when provided with the supporting evidence
(\autoref{exp}).
Together, these results support the distinction between
reading available evidence and reconstructing the authoritative
state needed to answer a question.

\paragraph{The bottleneck persists across question categories.}
State reconstruction dominates across the category groups
in \autoref{fig:error anaylsis}b.
Although their outputs differ, these tasks share a prerequisite:
identifying the operative state and its supporting updates.
Current and Past State questions require selecting the correct
state at the relevant point; Derived State questions require
reconstructing that state before computation; Change History
questions require retaining the accepted update sequence;
and Conflict/Refusal questions require checking a request
against the maintained state.
Consequently, a reconstruction failure can affect all five
categories rather than appearing only in questions explicitly
asking about update history.

\paragraph{Lost updates and stale values are the largest
reconstruction failure modes.}
We further classify state-reconstruction errors by their
dominant erroneous state operation
(\autoref{fig:error anaylsis}c).
\textbf{Lost/collapsed updates account for 30\% of all errors},
while \textbf{stale values account for 27.1\%}.
The former occurs when a governing update is omitted or
collapsed during reconstruction, so its effect is absent
from the maintained history or state.
The latter occurs when a superseded value remains in use
despite an operative replacement.
These subtypes distinguish failure to preserve an update
from failure to resolve which value it makes authoritative.
Their percentages use \emph{all incorrect responses} as the
denominator, not only state-reconstruction errors.

\paragraph{Reliable reconstruction requires multiple
state operations.}
The remaining subtypes span computation on the wrong state,
entity tracking, withdrawn values, authority resolution,
and update integration. Each accounts for approximately
10\% or less of all errors, but together they demonstrate
that reconstruction is not reducible to a single operation.
Retaining an update is insufficient if it is associated with
the wrong entity, treated as authoritative outside its scope,
or combined incorrectly with other changes.
Likewise, selecting the latest displayed value is insufficient
when a textual correction or withdrawal determines the
accepted state.

These findings complement the update-count analysis in
\autoref{fig:acc_haystack_update_recency}, where accuracy
generally declines as more governing updates must be integrated.
Together, the update-count trend and diverse error subtypes
suggest that reliable memory requires a state-aware approach:
\textbf{preserving and reconciling accumulating updates across
multiple state operations}.
This explains why DSV-Mem tests more than isolated visual
recognition or retrieval: successful answers depend on
maintaining the authoritative evolution of dense professional
artifacts.

\subsection{Additional Fine-Tuning Experiment}
\label{app:finetuning}

We explore whether supervised fine-tuning (SFT) on final-answer
labels improves DSV-Mem performance using Qwen3.5-9B.
We consider two settings: an in-domain 80/20 train--test split
and a cross-scenario split using eight scenarios for training
and two held-out scenarios for evaluation.
Training supervises only the gold answer, without intermediate
evidence-selection, state-reconstruction, or reasoning traces.

\begin{table}[t]
\centering
\caption{\textbf{Label-only fine-tuning on DSV-Mem.}
Accuracies are reported in percentages.
Category Avg.\ is the unweighted mean of the five categories.
ID denotes the in-domain split; OOD denotes the held-out-scenario
split.}
\label{tab:finetuning}
\resizebox{\linewidth}{!}{%
\begin{tabular}{llcccccc}
\toprule
\small
Setting & Model &
\shortstack{Current\\State} &
\shortstack{Past\\State} &
\shortstack{Derived\\State} &
\shortstack{Change\\History} &
\shortstack{Conflict/\\Refusal} &
\shortstack{Category\\Avg.} \\
\midrule
ID & Qwen3.5-9B
& 9.4 & 9.9 & 6.3 & 7.2 & 4.8 & 7.52 \\
ID & Qwen3.5-9B + SFT
& 11.2 & 8.5 & 6.4 & 8.2 & 26.6 & 12.18 \\
\midrule
OOD & Qwen3.5-9B
& 11.2 & 8.8 & 5.4 & 7.4 & 5.6 & 7.68 \\
OOD & Qwen3.5-9B + SFT
& 9.4 & 8.8 & 6.2 & 9.2 & 24.5 & 11.62 \\
\bottomrule
\end{tabular}%
}
\end{table}

\paragraph{Results.}
As shown in \autoref{tab:finetuning}, Conflict/Refusal improves
from 4.8\% to 26.6\% in-domain and from 5.6\% to 24.5\% on held-out scenarios,
suggesting that this category is more amenable to final-answer
supervision. In contrast, Current State, Past State,
Derived State, and Change History remain close to baseline and do not reflect broad improvement in state reconstruction.

\paragraph{Interpretation.}
Label-only supervision provides a target answer without
explicitly teaching the intermediate operations required to
obtain it: selecting relevant visual evidence, preserving
governing updates, and reconciling their cumulative effects
before answering. The limited gains outside Conflict/Refusal
are consistent with the state-reconstruction bottleneck
identified in \autoref{sec:error_analysis}.
These results suggest that final-answer SFT alone, under the
tested configuration, is insufficient to substantially improve
this capability; supervision targeting evidence selection
and state evolution remains a promising direction.

\subsection{Performance across scenarios}
\label{app:scenario_results}

\autoref{tab:gemini_scenario_results} reports Gemini 3.7 Flash
Medium performance across the ten scenarios. Mean accuracy ranges
from 18.0\% on incident-postmortem reconciliation to 81.0\% on
robotics evaluation, with eight scenarios below 50\%.
Although performance varies substantially across
scenarios, the aggregate difficulty is \emph{not} confined to a single
workflow. The scenario-level spread also highlights heterogeneous workflow difficulty;
its relationship to update count, artifact density, and authority-operation
mix remains an important direction for further analysis. Each scenario contains 100 questions with the same
category distribution, so the overall score equally weights
the ten scenarios. 

\begin{table}[t]
\centering
\small
\caption{Gemini 3.7 Flash Medium accuracy (\%) by scenario.
Results report mean $\pm$ sample standard deviation over three
runs on the same 100 questions per scenario.}
\label{tab:gemini_scenario_results}
\begin{tabular}{lc}
\toprule
Scenario & Accuracy (\%) \\
\midrule
Tax-return reconciliation       & $29.7 \pm 1.5$ \\
E-commerce reconciliation       & $26.3 \pm 2.5$ \\
ML agent research               & $37.0 \pm 2.6$ \\
Clinical-trial database lock    & $42.0 \pm 1.7$ \\
Sales-pipeline forecasting      & $29.7 \pm 2.9$ \\
SRE incident postmortem         & $18.0 \pm 2.0$ \\
Scientific-poster revision      & $40.0 \pm 1.0$ \\
Product-campaign design         & $36.0 \pm 4.0$ \\
Expo floor-plan revision        & $56.7 \pm 2.9$ \\
Robotics evaluation             & $81.0 \pm 1.0$ \\
\midrule
Overall                         & $39.6 \pm 0.8$ \\
\bottomrule
\end{tabular}
\end{table}

\subsection{Memory-management integrations}
\label{app:memory_integrations}

By design, these methods typically distill interaction histories
into compact memories and answer using retrieved memories at
inference time. Subsequent answers therefore depend on what
survives memory construction and retrieval. We supply historical
visual information during construction: SimpleMem and LightMem
receive detailed, question-independent transcriptions through
their text-based interfaces, while Mem0 and MIRIX receive images
directly through their multimodal ingestion interfaces. To reduce
transcription loss, the configured model is explicitly instructed
to preserve labels, values, statuses, spatial groupings, and
formatting cues. Each method then applies its own memory-construction
and retrieval procedures to the chronological conversation.

At answer time, the backbone receives retrieved textual memories
and any question-attached images. Historical visual information
is therefore mediated by constructed memories, rather than withheld
during ingestion. Their underperformance on DSV-Mem highlights
the challenge of preserving dense, state-bearing details through
this process. This also partially \emph{motivates} our self-ledger design: preserving
update-level records and source links allows fine-grained details
to be revisited rather than relying solely on compressed
representations. Memory construction is scenario-specific and
question-independent, using only visible evidence without gold
answers or hidden state graphs.
\subsection{Visible-evidence self-ledger}
\label{app:self_ledger}

We explore a training-free self-ledger that explicitly preserves
updates and their provenance. The model processes the conversation
in chronological windows, using visible dialogue and images,
image-derived OCR, and visual-difference hints. It appends
entity--field events recording observed values, changes,
source turns, and authority cues, while maintaining a compact
index of entity aliases, standing instructions, and unresolved
conflicts. Displayed values are distinguished from accepted
working values, allowing textual corrections or rejected visual
revisions to remain explicit without overwriting history.

The ledger is frozen before evaluation questions are introduced.
For each question, a retrieval step selects relevant events,
temporal scope, source dialogue, and original visual evidence.
The model uses this evidence to check an initial answer and
produce a ledger-assisted draft; a final arbitration step
reconciles the candidate answers against their visible sources.
Ledger construction receives no evaluation questions, gold
answers, or benchmark state graphs. This exploratory procedure
tests whether explicit update preservation and reconciliation
can improve performance without model training.




\begin{thebibliography}{38}
\providecommand{\natexlab}[1]{#1}
\providecommand{\url}[1]{\texttt{#1}}
\expandafter\ifx\csname urlstyle\endcsname\relax
  \providecommand{\doi}[1]{doi: #1}\else
  \providecommand{\doi}{doi: \begingroup \urlstyle{rm}\Url}\fi

\bibitem[Bei et~al.(2026)Bei, Wei, Ning, Zhao, Liu, Lin, Zhu, Hamann, He, and Tong]{bei2026memgallery}
Yuanchen Bei, Tianxin Wei, Xuying Ning, Yanjun Zhao, Zhining Liu, Xiao Lin, Yada Zhu, Hendrik Hamann, Jingrui He, and Hanghang Tong.
\newblock {Mem-Gallery}: Benchmarking multimodal long-term conversational memory for {MLLM} agents, 2026.
\newblock URL \url{https://arxiv.org/abs/2601.03515}.

\bibitem[Chhikara et~al.(2025)Chhikara, Khant, Aryan, Singh, and Yadav]{chhikara2025mem0}
Prateek Chhikara, Dev Khant, Saket Aryan, Taranjeet Singh, and Deshraj Yadav.
\newblock {Mem0}: Building production-ready {AI} agents with scalable long-term memory, 2025.
\newblock URL \url{https://arxiv.org/abs/2504.19413}.

\bibitem[Fang et~al.(2026)Fang, Deng, Xu, Jiang, Tang, Xu, Deng, Yao, Wang, Qiao, Chen, and Zhang]{fang2026lightmem}
Jizhan Fang, Xinle Deng, Haoming Xu, Ziyan Jiang, Yuqi Tang, Ziwen Xu, Shumin Deng, Yunzhi Yao, Mengru Wang, Shuofei Qiao, Huajun Chen, and Ningyu Zhang.
\newblock {LightMem}: Lightweight and efficient memory-augmented generation.
\newblock In \emph{International Conference on Learning Representations}, 2026.
\newblock URL \url{https://arxiv.org/abs/2510.18866}.

\bibitem[{Gemma Team}(2026)]{gemma4technicalreport2026}
{Gemma Team}.
\newblock {Gemma 4 Technical Report}, 2026.
\newblock URL \url{https://arxiv.org/abs/2607.02770}.

\bibitem[{Google DeepMind}(2026{\natexlab{a}})]{gemini35flashlite2026}
{Google DeepMind}.
\newblock {Gemini 3.5 Flash-Lite}, July 2026{\natexlab{a}}.
\newblock URL \url{https://deepmind.google/models/model-cards/gemini-3-5-flash-lite/}.
\newblock Model card.

\bibitem[{Google DeepMind}(2026{\natexlab{b}})]{gemini36flash2026}
{Google DeepMind}.
\newblock {Gemini 3.6 Flash}, July 2026{\natexlab{b}}.
\newblock URL \url{https://deepmind.google/models/model-cards/gemini-3-6-flash/}.
\newblock Model card.

\bibitem[{Google DeepMind}(2026{\natexlab{c}})]{google2026gemini31pro}
{Google DeepMind}.
\newblock {Gemini 3.1 Pro Model Card}.
\newblock Technical report, Google DeepMind, 2026{\natexlab{c}}.
\newblock URL \url{https://deepmind.google/models/model-cards/gemini-3-1-pro/}.

\bibitem[{Google DeepMind}(2026{\natexlab{d}})]{google2026gemini37flash}
{Google DeepMind}.
\newblock {Gemini 3.7 Flash}, August 2026{\natexlab{d}}.
\newblock URL \url{https://deepmind.google/models/model-cards/gemini-3-7-flash/}.
\newblock Model card.

\bibitem[Guo et~al.(2026)Guo, Jiao, Shi, Quan, Zhang, Li, Che, Xu, Liu, Liu, Kapadia, Pavlovic, Liu, Wang, Shi, Metaxas, and Tang]{guo2026memeye}
Minghao Guo, Qingyue Jiao, Zeru Shi, Yihao Quan, Boxuan Zhang, Danrui Li, Liwei Che, Wujiang Xu, Shilong Liu, Zirui Liu, Mubbasir Kapadia, Vladimir Pavlovic, Jiang Liu, Mengdi Wang, Yiyu Shi, Dimitris~N. Metaxas, and Ruixiang Tang.
\newblock {MemEye}: A visual-centric evaluation framework for multimodal agent memory, 2026.
\newblock URL \url{https://arxiv.org/abs/2605.15128}.

\bibitem[Hartley(1928)]{hartley1928transmission}
R.~V.~L. Hartley.
\newblock Transmission of information.
\newblock \emph{Bell System Technical Journal}, 7\penalty0 (3):\penalty0 535--563, 1928.
\newblock \doi{10.1002/j.1538-7305.1928.tb01236.x}.

\bibitem[Hu et~al.(2026)Hu, Li, Gao, Chen, Bai, Xu, Lin, Li, Han, Pei, and Deng]{hu2026evermembench}
Chuanrui Hu, Tong Li, Xingze Gao, Hongda Chen, Yi~Bai, Dannong Xu, Tianwei Lin, Xiaohong Li, Yunyun Han, Jian Pei, and Yafeng Deng.
\newblock Evaluating long-horizon memory for multi-party collaborative dialogues, 2026.
\newblock URL \url{https://arxiv.org/abs/2602.01313}.

\bibitem[Hu et~al.(2025)Hu, Xiong, Yi, Wei, Xiao, Chen, Ye, Tao, Zhou, Zhao, Li, Xu, Wang, Xu, Qiao, Wang, Kuang, Zeng, Wang, Li, Jiang, Zhou, Wang, Yin, Zhao, Yang, Wu, Zhang, and Wu]{hu2025osagents}
Xueyu Hu, Tao Xiong, Biao Yi, Zishu Wei, Ruixuan Xiao, Yurun Chen, Jiasheng Ye, Meiling Tao, Xiangxin Zhou, Ziyu Zhao, Yuhuai Li, Shengze Xu, Shenzhi Wang, Xinchen Xu, Shuofei Qiao, Zhaokai Wang, Kun Kuang, Tieyong Zeng, Liang Wang, Jiwei Li, Yuchen~Eleanor Jiang, Wangchunshu Zhou, Guoyin Wang, Keting Yin, Zhou Zhao, Hongxia Yang, Fan Wu, Shengyu Zhang, and Fei Wu.
\newblock {OS} agents: A survey on {MLLM}-based agents for computer, phone and browser use.
\newblock In \emph{Proceedings of the 63rd Annual Meeting of the Association for Computational Linguistics (Volume 1: Long Papers)}, pp.\  7436--7465. Association for Computational Linguistics, 2025.
\newblock \doi{10.18653/v1/2025.acl-long.369}.
\newblock URL \url{https://aclanthology.org/2025.acl-long.369/}.

\bibitem[Huang et~al.(2026)Huang, Liu, Tian, Chen, Chen, Wu, Zhang, Guo, and Zhou]{huang2026m3exam}
Zhengjun Huang, Wenxuan Liu, Zhoujin Tian, Wei Chen, Junle Chen, Yuqian Wu, Fangyuan Zhang, Qintian Guo, and Xiaofang Zhou.
\newblock {M$^3$Exam}: Benchmarking multimodal memory for realistic user-agent interactions, 2026.
\newblock URL \url{https://arxiv.org/abs/2606.07402}.

\bibitem[Jim{\'e}nez~Guti{\'e}rrez et~al.(2024)Jim{\'e}nez~Guti{\'e}rrez, Shu, Gu, Yasunaga, and Su]{gutierrez2025hipporag}
Bernal Jim{\'e}nez~Guti{\'e}rrez, Yiheng Shu, Yu~Gu, Michihiro Yasunaga, and Yu~Su.
\newblock {HippoRAG}: Neurobiologically inspired long-term memory for large language models.
\newblock In \emph{Advances in Neural Information Processing Systems}, volume~37, 2024.
\newblock URL \url{https://proceedings.neurips.cc/paper_files/paper/2024/hash/6ddc001d07ca4f319af96a3024f6dbd1-Abstract-Conference.html}.

\bibitem[Kang et~al.(2025)Kang, Ji, Zhao, and Bai]{kang2025memoryos}
Jiazheng Kang, Mingming Ji, Zhe Zhao, and Ting Bai.
\newblock Memory {OS} of {AI} agent.
\newblock In \emph{Proceedings of the 2025 Conference on Empirical Methods in Natural Language Processing}, pp.\  25961--25970. Association for Computational Linguistics, 2025.
\newblock \doi{10.18653/v1/2025.emnlp-main.1318}.
\newblock URL \url{https://aclanthology.org/2025.emnlp-main.1318/}.

\bibitem[{Kimi Team}(2026{\natexlab{a}})]{kimi2026k25}
{Kimi Team}.
\newblock {Kimi K2.5}: Visual agentic intelligence, 2026{\natexlab{a}}.
\newblock URL \url{https://arxiv.org/abs/2602.02276}.

\bibitem[{Kimi Team}(2026{\natexlab{b}})]{kimi2026k3}
{Kimi Team}.
\newblock {Kimi K3}: Open frontier intelligence, 2026{\natexlab{b}}.
\newblock URL \url{https://arxiv.org/abs/2607.24653}.

\bibitem[Liu et~al.(2026{\natexlab{a}})Liu, Yang, Pu, Liu, Long, Guo, Chen, Weng, Kochkina, Kaur, Smiley, Liu, Zou, Liu, Bu, Peng, and Wang]{liu2026worldmemarena}
Chengzhi Liu, Yuzhe Yang, Sophia~Xiao Pu, Yepeng Liu, Lin Long, Yichen Guo, Nuo Chen, Zhaotian Weng, Elena Kochkina, Simerjot Kaur, Charese Smiley, Xiaomo Liu, James Zou, Sheng Liu, Yuheng Bu, Songyou Peng, and Xin~Eric Wang.
\newblock {WorldMemArena}: Evaluating multimodal agent memory through action-world interaction, 2026{\natexlab{a}}.
\newblock URL \url{https://arxiv.org/abs/2605.29341}.

\bibitem[Liu et~al.(2026{\natexlab{b}})Liu, Su, Xia, Han, Zheng, Xie, Ding, and Yao]{liu2026simplemem}
Jiaqi Liu, Yaofeng Su, Peng Xia, Siwei Han, Zeyu Zheng, Cihang Xie, Mingyu Ding, and Huaxiu Yao.
\newblock {SimpleMem}: Efficient lifelong memory for {LLM} agents, 2026{\natexlab{b}}.
\newblock URL \url{https://arxiv.org/abs/2601.02553}.

\bibitem[Liu et~al.(2026{\natexlab{c}})Liu, Sun, Cheng, Lei, Chen, Wen, Yang, Fu, Cai, Deng, Yu, Hu, Shi, and Wang]{liu2026memverse}
Junming Liu, Yifei Sun, Weihua Cheng, Haodong Lei, Yirong Chen, Licheng Wen, Xuemeng Yang, Daocheng Fu, Pinlong Cai, Nianchen Deng, Yi~Yu, Shuyue Hu, Botian Shi, and Ding Wang.
\newblock {MemVerse}: Multimodal memory for lifelong learning agents, 2026{\natexlab{c}}.
\newblock URL \url{https://arxiv.org/abs/2512.03627v2}.
\newblock Version 2.

\bibitem[Maharana et~al.(2024)Maharana, Lee, Tulyakov, Bansal, Barbieri, and Fang]{maharana2024locomo}
Adyasha Maharana, Dong-Ho Lee, Sergey Tulyakov, Mohit Bansal, Francesco Barbieri, and Yuwei Fang.
\newblock Evaluating very long-term conversational memory of {LLM} agents.
\newblock In \emph{Proceedings of the 62nd Annual Meeting of the Association for Computational Linguistics (Volume 1: Long Papers)}, pp.\  13851--13870. Association for Computational Linguistics, 2024.
\newblock \doi{10.18653/v1/2024.acl-long.747}.
\newblock URL \url{https://aclanthology.org/2024.acl-long.747/}.

\bibitem[{MiniMax}(2026)]{minimax2026m3}
{MiniMax}.
\newblock {MiniMax-M3}, 2026.
\newblock URL \url{https://huggingface.co/MiniMaxAI/MiniMax-M3}.
\newblock Official model card.

\bibitem[Packer et~al.(2024)Packer, Wooders, Lin, Fang, Patil, Stoica, and Gonzalez]{packer2024memgpt}
Charles Packer, Sarah Wooders, Kevin Lin, Vivian Fang, Shishir~G. Patil, Ion Stoica, and Joseph~E. Gonzalez.
\newblock {MemGPT}: Towards {LLMs} as operating systems, 2024.
\newblock URL \url{https://arxiv.org/abs/2310.08560v2}.
\newblock Version 2.

\bibitem[Park et~al.(2023)Park, O'Brien, Cai, Morris, Liang, and Bernstein]{park2023generativeagents}
Joon~Sung Park, Joseph~C. O'Brien, Carrie~J. Cai, Meredith~Ringel Morris, Percy Liang, and Michael~S. Bernstein.
\newblock Generative agents: Interactive simulacra of human behavior.
\newblock In \emph{Proceedings of the 36th Annual ACM Symposium on User Interface Software and Technology}. Association for Computing Machinery, 2023.
\newblock \doi{10.1145/3586183.3606763}.
\newblock URL \url{https://arxiv.org/abs/2304.03442}.

\bibitem[{Qwen Team}(2026)]{qwen2026qwen35}
{Qwen Team}.
\newblock {Qwen3.5: Towards Native Multimodal Agents}, 2026.
\newblock URL \url{https://qwen.ai/blog?id=qwen3.5}.

\bibitem[Ren et~al.(2026)Ren, Wang, Du, Xie, Liu, Yang, Feng, Pan, Zheng, Xu, Li, Song, Wong, and See]{ren2026memlens}
Xiyu Ren, Zhaowei Wang, Yiming Du, Zhongwei Xie, Chi Liu, Xinlin Yang, Haoyue Feng, Wenjun Pan, Tianshi Zheng, Baixuan Xu, Zhengnan Li, Yangqiu Song, Ginny Wong, and Simon See.
\newblock {MemLens}: Benchmarking multimodal long-term memory in large vision-language models, 2026.
\newblock URL \url{https://arxiv.org/abs/2605.14906}.

\bibitem[Wang et~al.(2025)Wang, Zou, Song, Feng, Fang, Lu, Liu, Luo, Liang, Huang, et~al.]{wang2025uitars}
Haoming Wang, Haoyang Zou, Huatong Song, Jiazhan Feng, Junjie Fang, Junting Lu, Longxiang Liu, Qinyu Luo, Shihao Liang, Shijue Huang, et~al.
\newblock {UI-TARS-2} technical report: Advancing {GUI} agent with multi-turn reinforcement learning.
\newblock \emph{arXiv preprint arXiv:2509.02544}, 2025.
\newblock URL \url{https://arxiv.org/abs/2509.02544}.

\bibitem[Wang et~al.(2023)Wang, Dong, Cheng, Liu, Yan, Gao, and Wei]{wang2023longmem}
Weizhi Wang, Li~Dong, Hao Cheng, Xiaodong Liu, Xifeng Yan, Jianfeng Gao, and Furu Wei.
\newblock Augmenting language models with long-term memory.
\newblock In \emph{Advances in Neural Information Processing Systems}, volume~36, 2023.
\newblock URL \url{https://proceedings.neurips.cc/paper_files/paper/2023/hash/ebd82705f44793b6f9ade5a669d0f0bf-Abstract-Conference.html}.

\bibitem[Wang \& Chen(2025)Wang and Chen]{wang2025mirix}
Yu~Wang and Xi~Chen.
\newblock {MIRIX}: Multi-agent memory system for {LLM}-based agents, 2025.
\newblock URL \url{https://arxiv.org/abs/2507.07957}.

\bibitem[Wei et~al.(2022)Wei, Wang, Schuurmans, Bosma, Ichter, Xia, Chi, Le, and Zhou]{wei2022chain}
Jason Wei, Xuezhi Wang, Dale Schuurmans, Maarten Bosma, Brian Ichter, Fei Xia, Ed~H. Chi, Quoc~V. Le, and Denny Zhou.
\newblock Chain-of-thought prompting elicits reasoning in large language models.
\newblock In \emph{Advances in Neural Information Processing Systems}, volume~35, pp.\  24824--24837, 2022.
\newblock URL \url{https://proceedings.neurips.cc/paper_files/paper/2022/hash/9d5609613524ecf4f15af0f7b31abca4-Abstract-Conference.html}.

\bibitem[Wu et~al.(2025)Wu, Wang, Yu, Zhang, Chang, and Yu]{wu2025longmemeval}
Di~Wu, Hongwei Wang, Wenhao Yu, Yuwei Zhang, Kai-Wei Chang, and Dong Yu.
\newblock {LongMemEval}: Benchmarking chat assistants on long-term interactive memory.
\newblock In \emph{International Conference on Learning Representations}, 2025.
\newblock URL \url{https://arxiv.org/abs/2410.10813}.

\bibitem[Wu et~al.(2026)Wu, Ji, Kawatkar, Kwan, Gu, Peng, and Chang]{wu2026longmemevalv2}
Di~Wu, Zixiang Ji, Asmi Kawatkar, Bryan Kwan, Jia-Chen Gu, Nanyun Peng, and Kai-Wei Chang.
\newblock {LongMemEval-V2}: Evaluating long-term agent memory toward experienced colleagues, 2026.
\newblock URL \url{https://arxiv.org/abs/2605.12493}.

\bibitem[Xi et~al.(2023)Xi, Chen, Guo, He, Ding, Hong, Zhang, Wang, Jin, Zhou, et~al.]{xi2023rise}
Zhiheng Xi, Wenxiang Chen, Xin Guo, Wei He, Yiwen Ding, Boyang Hong, Ming Zhang, Junzhe Wang, Senjie Jin, Enyu Zhou, et~al.
\newblock The rise and potential of large language model based agents: A survey.
\newblock \emph{arXiv preprint arXiv:2309.07864}, 2023.
\newblock URL \url{https://arxiv.org/abs/2309.07864}.

\bibitem[Xu et~al.(2025)Xu, Liang, Mei, Gao, Tan, and Zhang]{xu2025amem}
Wujiang Xu, Zujie Liang, Kai Mei, Hang Gao, Juntao Tan, and Yongfeng Zhang.
\newblock {A-Mem}: Agentic memory for {LLM} agents.
\newblock In \emph{Advances in Neural Information Processing Systems}, volume~38, 2025.
\newblock URL \url{https://proceedings.neurips.cc/paper_files/paper/2025/hash/19909c36f51abc4856b4560aff3d36d6-Abstract-Conference.html}.

\bibitem[{Z.ai}(2025)]{zai2025glm46v}
{Z.ai}.
\newblock {GLM-4.6V}, 2025.
\newblock URL \url{https://huggingface.co/zai-org/GLM-4.6V}.
\newblock Official model card.

\bibitem[Zheng et~al.(2023)Zheng, Chiang, Sheng, Zhuang, Wu, Zhuang, Lin, Li, Li, Xing, Zhang, Gonzalez, and Stoica]{zheng2023judging}
Lianmin Zheng, Wei-Lin Chiang, Ying Sheng, Siyuan Zhuang, Zhanghao Wu, Yonghao Zhuang, Zi~Lin, Zhuohan Li, Dacheng Li, Eric~P. Xing, Hao Zhang, Joseph~E. Gonzalez, and Ion Stoica.
\newblock Judging {LLM-as-a-Judge} with {MT-Bench} and {Chatbot Arena}.
\newblock In \emph{Advances in Neural Information Processing Systems}, volume~36, 2023.
\newblock URL \url{https://proceedings.neurips.cc/paper_files/paper/2023/hash/91f18a1287b398d378ef22505bf41832-Abstract-Datasets_and_Benchmarks.html}.

\bibitem[Zhong et~al.(2023)Zhong, Guo, Gao, Ye, and Wang]{zhong2023memorybank}
Wanjun Zhong, Lianghong Guo, Qiqi Gao, He~Ye, and Yanlin Wang.
\newblock {MemoryBank}: Enhancing large language models with long-term memory, 2023.
\newblock URL \url{https://arxiv.org/abs/2305.10250}.

\bibitem[Zhu et~al.(2026)Zhu, Yang, Wang, Shen, Guo, and Yang]{zhu2026h2hmem}
Shiping Zhu, Yibo Yang, Zhengyang Wang, Tiancheng Shen, Dandan Guo, and Ming-Hsuan Yang.
\newblock {H2HMem}: A multimodal memory benchmark for agents in human--human interactions, 2026.
\newblock URL \url{https://arxiv.org/abs/2606.09461}.

\end{thebibliography}
\end{document}